\documentclass[pdflatex,sn-mathphys]{sn-jnl}

\jyear{2022}%

\theoremstyle{thmstyleone}%
\theoremstyle{thmstyletwo}%

\theoremstyle{thmstylethree}%

\usepackage{color}

\usepackage{float}
\usepackage{etoolbox}
\makeatletter
\patchcmd{\ps@headings}
{\hbox to \hsize{\hfill Springer Nature 2021 \LaTeX\ template\hfill}}
{\hbox to \hsize{}}
{}
{}
\patchcmd{\ps@headings}
{\hbox to \hsize{\hfill Springer Nature 2021 \LaTeX\ template\hfill}}
{\hbox to \hsize{}}
{}
{}
\patchcmd{\ps@titlepage}
{\hbox to \hsize{\hfill Springer Nature 2021 \LaTeX\ template\hfill}}
{\hbox to \hsize{}}
{}
{}
\makeatother

\begin{document}
\small

\title[ ]{\Large Dual adaptive training of photonic neural networks}


\author[1,2]{\normalsize \fnm{Ziyang} \sur{Zheng}}
\equalcont{\normalsize These authors contributed equally to this work.}

\author[1]{\normalsize \fnm{Zhengyang} \sur{Duan}}
\equalcont{\normalsize These authors contributed equally to this work.}

\author[1]{\normalsize \fnm{Hang} \sur{Chen}}

\author[2]{\normalsize \fnm{Rui} \sur{Yang}}

\author[1]{\normalsize \fnm{Sheng} \sur{Gao}}

\author[1]{\normalsize \fnm{Haiou} \sur{Zhang}}

\author*[2]{\normalsize \fnm{Hongkai} \sur{Xiong}}

\author*[1,3]{\normalsize \fnm{Xing} \sur{Lin}}

\affil[1]{\orgdiv{\normalsize Department of Electronic Engineering}, \orgname{Tsinghua University}, \orgaddress{\city{Beijing}, \postcode{100084}, \country{China}}}

\affil[2]{\orgdiv{\normalsize Department of Electronic Engineering}, \orgname{Shanghai Jiao Tong University}, \orgaddress{\city{Shanghai}, \postcode{200240}, \country{China}}}

\affil[3]{\orgdiv{\normalsize Beijing National Research Center for Information Science and Technology}, \orgname{Tsinghua University}, \orgaddress{\city{Beijing}, \postcode{100084}, \country{China}}}

\email{\normalsize lin-x@tsinghua.edu.cn; xionghongkai@sjtu.edu.cn}


\abstract{Photonic neural network (PNN) is a remarkable analog artificial intelligence (AI) accelerator that computes with photons instead of electrons to feature low latency, high energy efficiency, and high parallelism. However, the existing training approaches cannot address the extensive accumulation of systematic errors in large-scale PNNs, resulting in a significant decrease in model performance in physical systems. Here, we propose dual adaptive training (DAT) that allows the PNN model to adapt to substantial systematic errors and preserves its performance during the deployment. By introducing the systematic error prediction networks with task-similarity joint optimization, DAT achieves the high similarity mapping between the PNN numerical models and physical systems and high-accurate gradient calculations during the dual backpropagation training. We validated the effectiveness of DAT by using diffractive PNNs and interference-based PNNs on image classification tasks. DAT successfully trained large-scale PNNs under major systematic errors and preserved the model classification accuracies comparable to error-free systems. The results further demonstrated its superior performance over the state-of-the-art in situ training approaches. DAT provides critical support for constructing large-scale PNNs to achieve advanced architectures and can be generalized to other types of AI systems with analog computing errors.}




\maketitle

\section{Main}\label{Main}

Artificial intelligence (AI), powered by deep neural networks (DNNs), utilizes brain-inspired information processing mechanisms to approach human-level performance in complex tasks~\cite{lecun:2015}, which has already achieved major applications ranging from translating languages~\cite{bahdanau:2015}, image recognition~\cite{rawat:2017}, cancer diagnosis~\cite{capper:2018} to fundamental science~\cite{torlai:2018}. The vast majority of AI algorithms have been implemented via digital electronic computing platforms, such as graphics- and tensor-processing units, to support their major computing power requirement. However, the requirements of AI for processors' computing performance have grown rapidly, greatly exceeding the development of digital electronic computing imposed by Moore's law and the upper limit of computing energy efficiency~\cite{xu:2022, patterson:2021, shainline:2017}. Constructing the photonic neural network (PNN) systems for AI tasks with analog photonic computing has attracted increasing attention and is expected to be the next-generation AI computing modality with the advantages of low latency, high bandwidth, and low power consumption. The fundamental characteristic of photons and principle of light-matter interactions, such as diffraction~\cite{lin:2018, yan:2019, zhou:2021} and interference~\cite{shen:2017, hughes:2019, williamson:2019} based on free-space optics or integrated photonic circuits, have been utilized to implement various neuromorphic photonic computing architectures, including convolutional neural networks~\cite{xu:2021, feldmann:2021, chang:2018, miscuglio:2020}, spiking neural networks~\cite{feldmann:2019, hamerly:2019, chakraborty:2019}, recurrent neural networks~\cite{hughes:2019_2, bueno:2018}, and reservoir computing~\cite{van:2017, larger:2017, brunner:2018}.

The effective training approach is one of the most critical aspects for DNNs to learn the reliable model and guarantee high inference accuracy. The DNNs constructed using software on a digital electronic computer generally train using backpropagation algorithm~\cite{lecun:1998}. Such training mechanism provides the basis for the \emph{in silico} training of photonic DNNs, which establishes the PNN models in computer to simulate physical systems, train models through backpropagation, and deploy the trained model parameters to physical systems. However, the inherent systematic errors of analog computing from different sources, e.g., geometric error and fabrication error, causes the deviation between the \emph{in silico} trained PNN model and physical system and results in the performance degeneration during the directly deploying~\cite{zhou:2021, zuo:2019, wright:2022}. To address the systematic errors, the \emph{in situ} training approaches, training PNNs on the physical systems with experimental measurements, have drawn increasing attention for optimizing the PNN models for practical applications~\cite{hughes:2018, zhou:2020, zhou:2021, filipovich:2021, wright:2022, spall:2022}. Nevertheless, the existing \emph{in situ} training methods still confront great challenges in training large-scale PNNs with major systematic errors, which hinder the construction of advanced architectures and limit the model performance in performing complex AI tasks. The reasons for this are mainly due to the inaccurate gradient calculations during the backpropagation caused by the imprecise modeling of PNN physical systems~\cite{filipovich:2021, wright:2022, spall:2022}, the requirement of extensive system measurements with layer-by-layer training processes~\cite{zhou:2021}, or the additional hardware configurations for backward optical field propagation~\cite{hughes:2018, zhou:2020}.

\begin{figure*}[!t]
    \centering
    \includegraphics[width=1.0\textwidth]{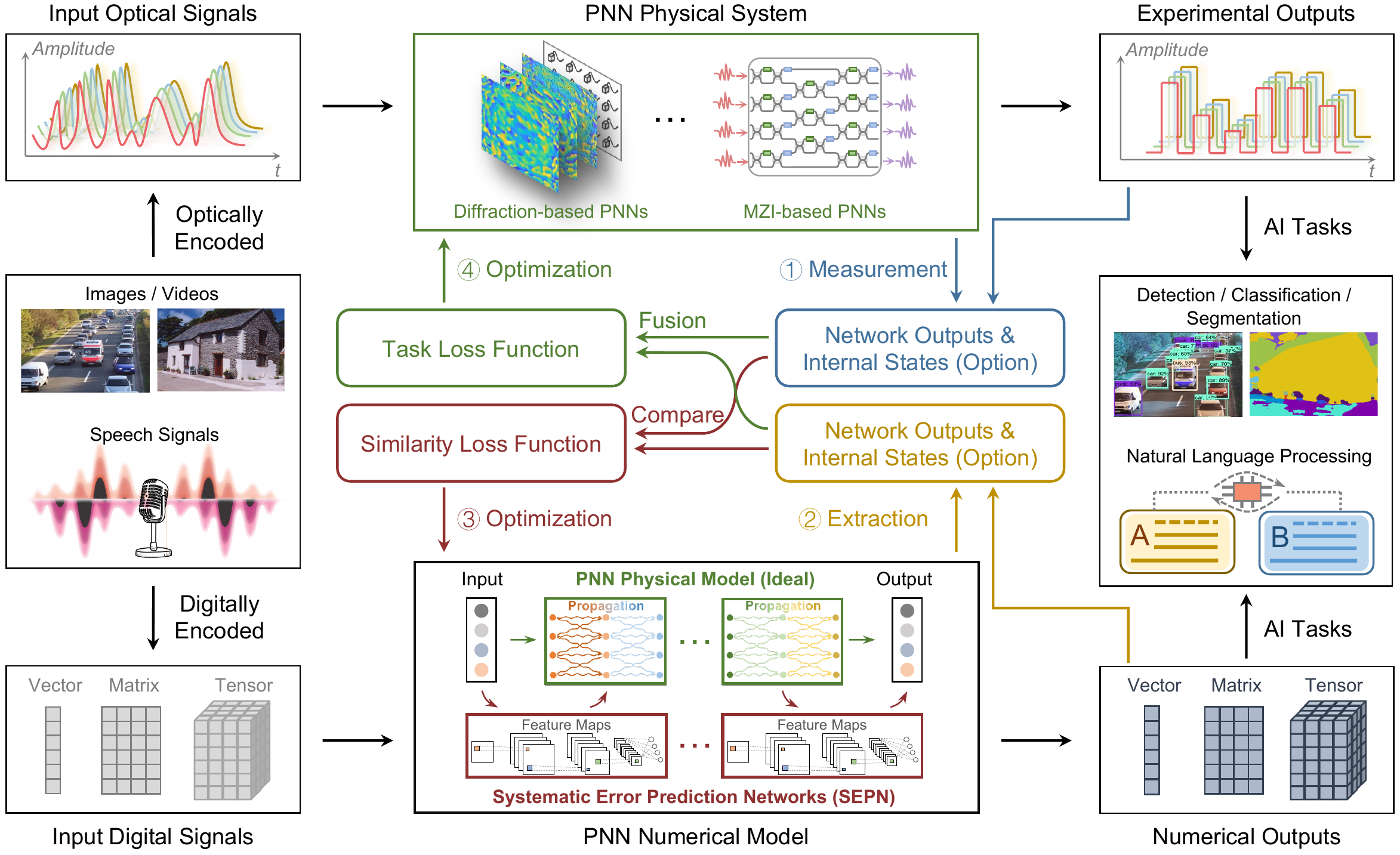}
    \caption{Training PNNs with DAT. The input information is optically and digitally encoded to feed into the PNN physical system and numerical model, respectively. To adapt to the systematic errors, each DAT training cycle consists of four steps to perform the dual backpropagation, which iteratively updates the PNN physical model and SEPN parameters by optimizing the task and similarity loss functions, respectively.}
    \label{DAT_procedure}
\end{figure*}

In this work, we propose dual adaptive training (DAT) for the end-to-end dual backpropagation training of large-scale PNNs, allowing the models to adapt to significant systematic errors without additional hardware configurations for backward optical field propagation. The basic principle of DAT for training PNNs with systematic errors is illustrated in Fig.~\ref{DAT_procedure}. To precisely model the PNN physical system, we introduce the systematic error prediction networks (SEPNs) in addition to the PNN physical model and develop the task-similarity joint optimization approach for dual backpropagation training. The DAT iteratively updates the network parameters of PNN and SEPNs in an end-to-end form for each input training sample. With the training of SEPNs to characterize the inherent systematic errors, the DAT establishes high similarity mapping between the PNN numerical models and physical systems, leading to high-accurate gradient calculation for PNN training. Each training sample is optically and digitally encoded as the input to the PNN physical system and forward numerical model, respectively. The physically measured and the numerically extracted network outputs are fused to obtain the task loss function and compared to obtain the similarity loss function. The network's internal states, if provided, can be further used to boost the DAT performance. The dual backpropagation training process of DAT minimizes the task and similarity loss functions to update the network parameters of the physical model and SEPNs, respectively, by calculating the gradients of the PNN numerical model. After the training, the PNN physical model, deployed on the physical system, can adapt to significant systematic errors from various sources. Therefore, DAT supports large-scale PNN training and mitigates the requirement of high-precision fabrication and system configurations.

The constructed PNN numerical model in a digital computer comprises the ideal PNN physical model and SEPNs for modeling the photonic computing process and inherent systematic errors, respectively. To facilitate learning the systematic errors of PNN layers, SEPNs are incorporated in the manner of residual connections for the PNN layers (see Fig.~\ref{DAT_procedure} and Methods), inspired by the residual neural networks proposed in~\cite{he:2016}. In this work, each SEPN module is configured with a complex-valued mini-UNet~\cite{ronneberger:2015} to guarantee its learning capacity for fitting the systematic errors of PNN layers. The target is to eliminate performance degradation while deploying PNN physical model parameters to the physical system. With the established PNN numerical model, the DAT for dual backpropagation training of the PNN numerical model consists of four main steps for each input training sample, including (1) the measurement of network outputs and optional internal states from the physical system; (2) the extraction of corresponding network outputs and optional internal states from the numerical model; (3) minimizing the similarity loss for backpropagation that updates the network parameters of SEPNs by comparing between the corresponding physical and numerical network's outputs and internal states; (4) minimizing the task loss for backpropagation that updates the network parameters of PNN physical model by replacing the numerical network's outputs and internal states with the physical measurements. We detail and formulate each of the training steps in Fig.~\ref{DAT_procedure} as follows:

First, each training sample is optically encoded and input to the PNN physical system to perform the forward inference, with which we obtain the physical network output $\textbf{P}_N$ for a $N$-layer neural network with input $\textbf{I}$. To further improve the training performance, the network internal states $\{\textbf{P}_1, \textbf{P}_2, ..., \textbf{P}_{N-1}\}$ can be measured at each layer's output. All the observations $\{\textbf{P}_n\}_{n=1}^{N}$ are set to be intensity, i.e., the absolute square of output complex optical fields, for facilitating the measurement. The optional internal states can boost the PNN training performance, especially under more severe systematic errors, but cause the additional cost of measurements. In practice, we can selectively measure a certain amount of internal states to reduce the number of measurements.

Second, the same training sample is also digitally encoded and input to the PNN numerical model to extract the internal states $\{\textbf{S}_1, \textbf{S}_2, ..., \textbf{S}_{N-1}\}$ and final observation $\textbf{S}_{N}$. Different from the counterpart physical measurements $\textbf{P}_n$, we set $\textbf{S}_n = \vert \textbf{S}_n \vert \exp(j\mathbf{\Phi}_{\textbf{S}_n})$ to be the complex optical fields with the amplitude $\vert \textbf{S}_n \vert$ and phase $\mathbf{\Phi}_{\textbf{S}_n}$ for facilitating the formulation of DAT process, which can be easily obtained during the numerically modeling of PNN forward inference. Ideally, the $ \vert \textbf{S}_n \vert ^2 = \textbf{P}_n $ if the systematic errors can be perfectly characterized with SEPNs. 

Third, we optimize the SEPNs' parameters of the PNN numerical model by minimizing the similarity loss function $L_{\mathrm{s}}$ as follows:
\begin{equation}\label{mse_loss}
    \min_{\Lambda} \left\{L_{\mathrm{s}}(\textbf{P}, \vert \textbf{S} \vert ^2) = \sum_{n=1}^N \alpha_n l_{\mathrm{mse}}(\textbf{P}_n, \vert \textbf{S}_n \vert ^2)=\sum_{n=1}^N \alpha_n \|\textbf{P}_n -  \vert \textbf{S}_n\vert ^2 \|_2^2 \right\},
\end{equation}
where $\textbf{P}=\{\textbf{P}_n\}_{n=1}^{N}$; $\textbf{S}=\{\textbf{S}_n\}_{n=1}^{N}$; $\Lambda$ refers to the learnable parameters of SEPNs; $l_{\mathrm{mse}}(\cdot)$ denotes a mean square error (MSE) function; $\alpha_n$ is the coefficient to weight the $n$-th MSE function. For each training sample, the parameters of the PNN physical model are fixed, and the gradients of $L_{\mathrm{s}}$ with respect to $\Lambda$ are calculated during the backpropagation to update SEPNs' parameters for one step. The optimization in Eq.~\eqref{mse_loss} aims to train SEPNs to minimize the deviation between the physically measured and numerically extracted network output and internal states for accurately modeling the PNN physical system.
We term the aforementioned training step for SEPNs as unitary mode since all SEPNs' parameters are optimized with a unitary loss function. In addition, the gradient calculation for SEPN modules can be implemented with separable mode (see Methods) when measuring internal states, where all SEPN modules are separated into several groups and optimized independently from each other. 

Fourth, we optimize the physical parameters of the PNN numerical model and deploy them to the physical system by minimizing the following task loss $L_{\mathrm{t}}$:
\begin{equation}\label{task_loss}
    \min_{\Omega} \left\{L_{\mathrm{t}}(\vert F_N(\textbf{P}_N, \textbf{S}_N)\vert^2, \textbf{T})  \right\},
\end{equation}
where $\Omega$ refers to the learnable parameters of the physical model; $\textbf{T}$ denotes the desired output; $L_{\mathrm{t}}$ is defined based on the target task implemented by PNNs, which is set to be the cross-entropy loss~\cite{lecun:2015} for classification tasks in this work; and $F_N(\textbf{P}_N, \textbf{S}_N)$ represents the output of the fusion function $F_N$ that replaces the amplitude of the numerically extracted network output with the physically measured counterpart. Furthermore, such fusion processes are applied for not only the network output but also the network internal states to maintain the interactions with the physical system; therefore, we have $\{F_n(\textbf{P}_n, \textbf{S}_n) = \sqrt{\textbf{P}_n} \exp(j\mathbf{\Phi}_{\textbf{S}_n})\}_{n=1}^N$. During the backpropagation for updating physical parameters with one step for each training sample, the parameters of SEPNs are fixed, and the fused network output and internal states are used for calculating the gradients of $L_{\mathrm{t}}$ with respect to the physical system parameters $\Omega$. The optimization in Eq.~\eqref{task_loss} aims to train the PNN physical model under systematic errors so that the PNN physical system deployed with physical parameters $\Omega$ can perform the target tasks.

The above training steps are repeated over all training samples to minimize the loss functions until convergence for obtaining the PNN numerical model and physical parameters $\Omega$ for the physical system. We term the training process as dual backpropagation training since the gradient calculation for updating the parameters of the PNN physical model and SEPNs rely on each other. Furthermore, the training of the PNN physical model promotes the training of SEPNs and vice versa. On the one hand, the optimization of physical parameters facilitates characterizing inherent systematic errors with SEPNs for task-specific physical models. On the other hand, the optimization of SEPNs' parameters facilitates performing the inference tasks with physical models under practical systematic errors. Besides, the state and output fusion processes allow the PNN physical model to further adapt the systematic errors and accelerate the convergence, especially when the SEPNs haven't fully characterized the systematic errors during the optimization. These underlying mechanisms guarantee the effectiveness and convergence of the proposed DAT.

We validate the effectiveness of DAT by applying it for training large-scale diffractive PNNs (DPNNs)~\cite{lin:2018, zhou:2021} and interference-based PNNs (MPNNs)~\cite{shen:2017, williamson:2019} under various systematic errors. The network settings and training processes for two types of models are detailed in the Methods section. Two benchmark datasets, i.e., the Modified National Institute of Standards and Technology (MNIST)~\cite{lecun:1998} and Fashion-MNIST (FMNIST)~\cite{xiao:2017}, were utilized for the performance evaluations. The results demonstrate the superior performance of DAT over the \emph{in silico} training with direct deployment and the state-of-the-art \emph{in situ} training method using physics-aware training (PAT)~\cite{wright:2022, spall:2022}.

\section{Results}\label{results}

\begin{figure*}[!ht]
    \renewcommand{\baselinestretch}{1.0}
    \centering
    \includegraphics[width=1.0\textwidth]{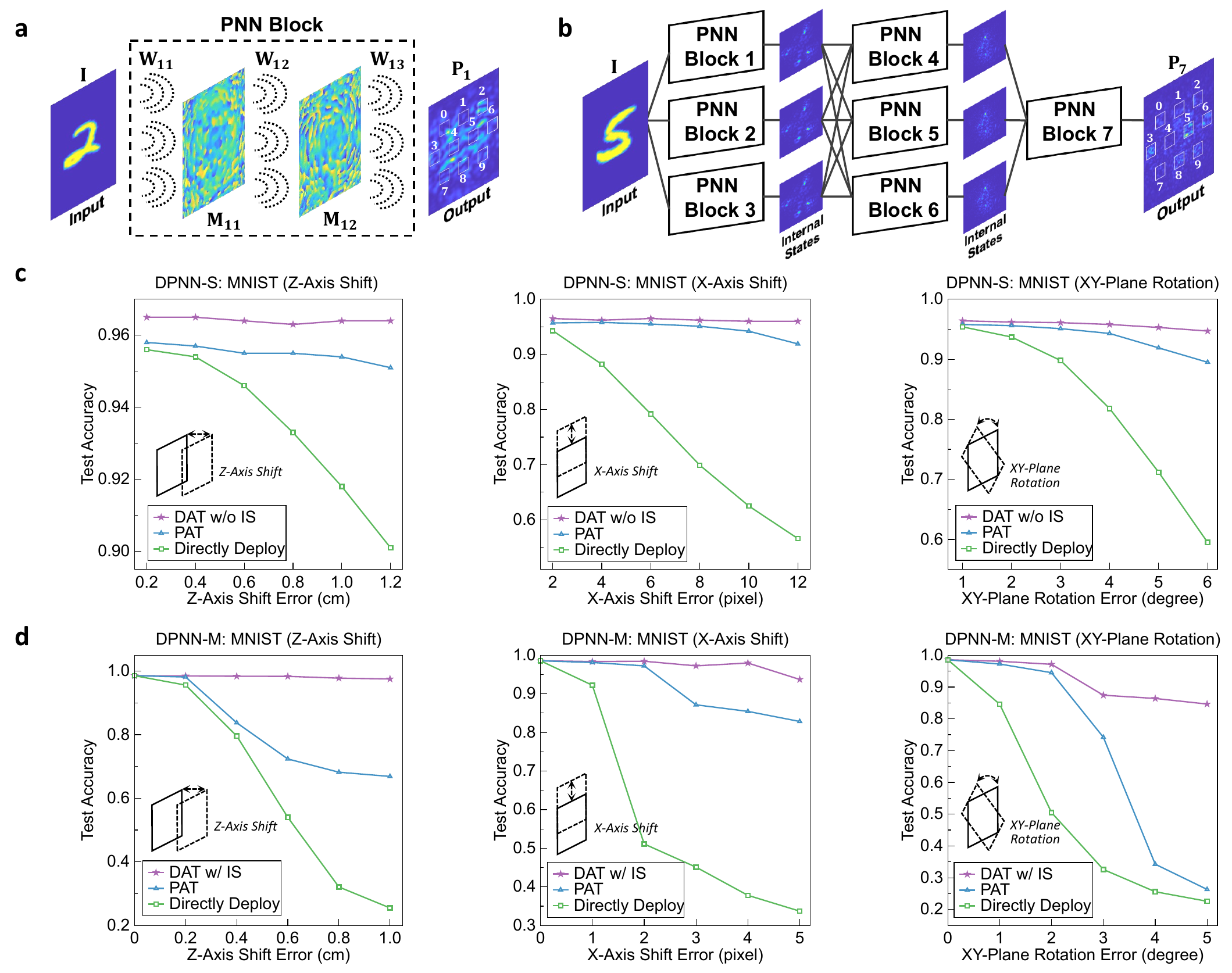}
    \caption{\textbf{Training DPNNs under three types of systematic errors for the MNIST classification.} $\textbf{a}$, DPNN-S with a single block. The block consists of two phase modulation layers and one intensity measurement layer, where the optical diffraction for weighted matrices occurs between adjacent layers. $\textbf{b}$, DPNN-M with seven hierarchically interconnected blocks, where each block is the same as the counterpart in $\textbf{a}$. The evaluations of DAT performance with DPNN-S ($\textbf{c}$) and DPNN-M ($\textbf{d}$). The performances of DAT are compared with the PAT and direct deployment of \emph{in silico} trained models under different amounts of systematic errors. The DPNN-S and DPNN-M are trained without and with the internal states (IS), respectively.}
    \label{dpnn_exp}
\end{figure*}

\subsection{Training DPNN with DAT}
\label{Experiments_diffractive_onns}

We built two types of DPNN architectures, i.e., the DPNN-S and DPNN-M, as illustrated in Fig.~\ref{dpnn_exp}a and Fig.~\ref{dpnn_exp}b, respectively. DPNN-S in Fig.~\ref{dpnn_exp}a was constructed using a single PNN block, where the block comprised the cascading of two phase modulation layers with transformation matrices $\textbf{M}_{11}, \textbf{M}_{12}$, followed by an optoelectronic intensity measurement layer at the output plane. The output layer of DPNN-S records the intensity $\textbf{P}_1$ of output optical fields for the input $\textbf{I}$. Specifically, the diffractive elements on a phase modulation layer are able to modulate the phase of input optical fields, and the secondary wave sources are generated via optical diffraction to interconnect to the next phase modulation layer or the output plane for intensity measurements. The forward propagation of a single PNN block has three free-space diffraction processes with three diffractive matrices $\textbf{W}_{11}, \textbf{W}_{12}$, and $\textbf{W}_{13}$. Therefore, the mathematical forward model of DPNN-S can be defined as: $\textbf{P}_1=\vert \textbf{W}_{13}\textbf{M}_{12}\textbf{W}_{12}\textbf{M}_{11}\textbf{W}_{11}\textbf{I} \vert^2$. To further demonstrate the effectiveness of the proposed method on DPNNs with larger network scales, we constructed DPNN-M~\cite{zhou:2021} that was designed with multiple PNN blocks to constitute multi-channel diffractive hidden layers with hierarchically interconnected structures. Each PNN block of DPNN-M is the same as the counterpart in DPNN-S, but their parameters are independent and not shared. DPNN-M has been demonstrated to achieve higher model performance yet inevitably accumulates more extensive systematic errors layer by layer with more complicated network structures.

For both DPNN-S and DPNN-M, the phase modulation coefficients are set as the learnable parameters and thus be optimized via end-to-end network training. In addition to recording intensity, the optoelectronic detectors on the output plane can be regarded as complex activation functions to accomplish the nonlinearity. The final output intensity is used for approximating the desired target by minimizing the specific task loss $L_{\mathrm{t}}$. We detail the DPNN settings and its training process with DAT in Methods. We further provide the pseudo-code of DAT in Supplementary Appendix A and Supplementary Algorithm S1, and elaborate on the standard training process in Supplementary Appendix B. For simplicity, all SEPN modules share the same network architecture for DPNN-S and DPNN-M (see Methods and Extended Data Fig.~1). Each SEPN was constructed as a complex-valued mini-UNet~\cite{ronneberger:2015} to extract hierarchical features, which is much simpler and lighter than a standard UNet. The trainable parameters of a SEPN module and UNet are 26,800 and 7,765,442, respectively, with a parameter ratio of 0.345\%.

We trained DPNN-S and DPNN-M with DAT for the MNIST and FMNIST classification tasks and compared their performance with PAT and direct deployment by considering four types of systematic errors in practical systems, i.e.,  \emph{Z-Axis} shift error, \emph{X-Axis} shift error, \emph{XY-Plane} rotation error, and phase shift error (see Methods for detail description). The first three errors are geometric errors mainly due to the imprecision of alignments, which are included in a layer-by-layer manner, each with the same amount of errors. For example, a single pixel \emph{X-Axis} shift error between successive layers in Fig.~\ref{dpnn_exp} results in the \emph{X-Axis} shift of three pixels in DPNN-S with a single block and nine pixels in DPNN-M. The phase shift error, modeled with a normal distribution with zero mean and standard deviation $\sigma$, is mainly caused by the imperfection of phase modulation devices that leads to the deviation of phase modulations. The classification performances of DPNN models were evaluated under individual and joint systematic errors to validate the effectiveness of DAT in various scenarios with different systematic error configurations.

\begin{figure*}[!t]
    \centering
    \includegraphics[width=1.0\textwidth]{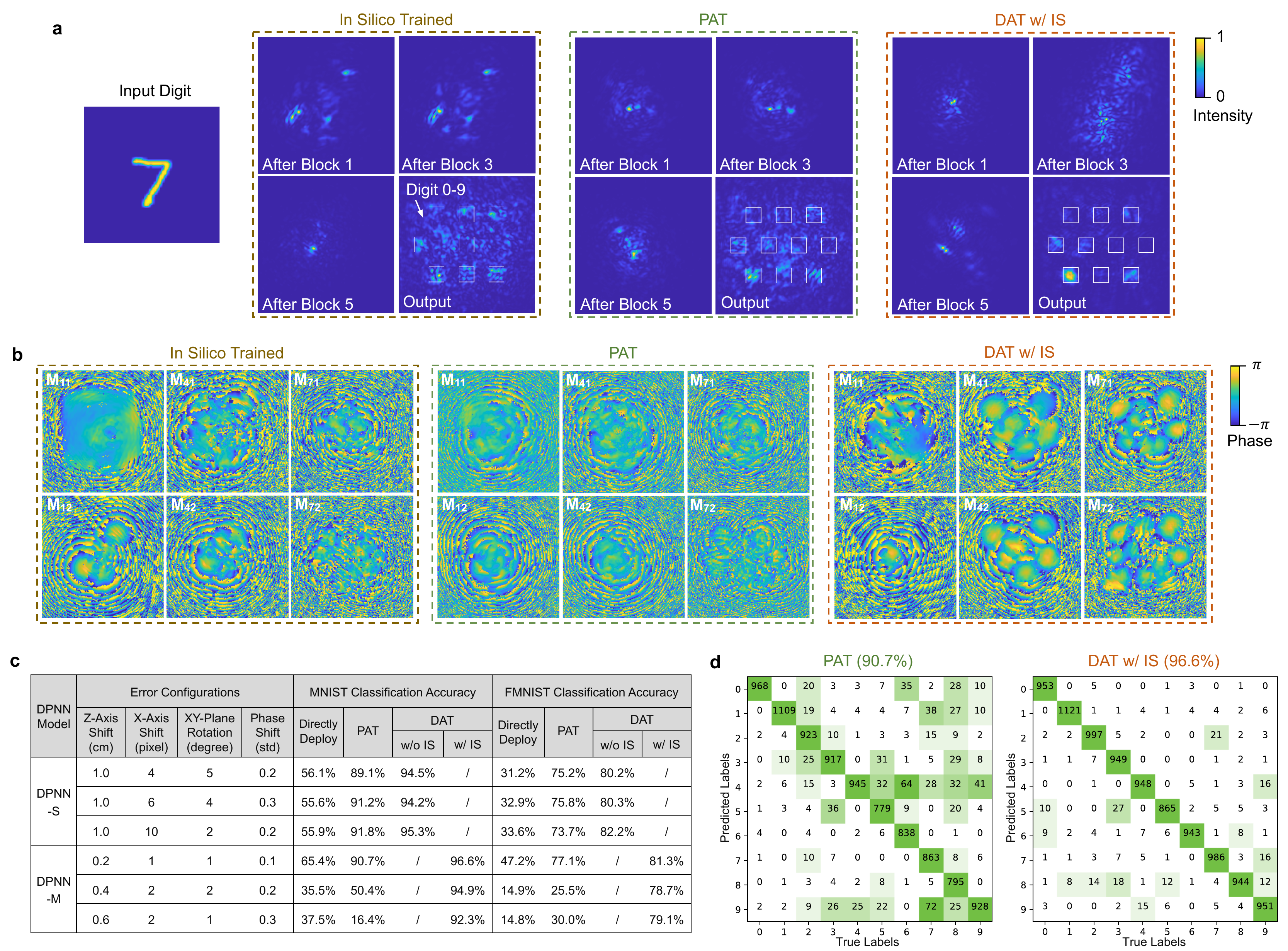}
    \caption{\textbf{Training DPNN under joint systematic errors for the MNIST and FMNIST classification.} The performance of DAT is evaluated on the DPNN-S and DPNN-M architectures and compared with the PAT and direct deployment of \emph{in silico} trained model under different joint systematic error configurations, as shown in Table $\textbf{c}$. The first configuration of DPNN-M listed in Table~$\textbf{c}$ was selected for the visualization of the network internal states, phase modulation layers, and confusion matrices on the MNIST classification in $\textbf{a}$, $\textbf{b}$, and $\textbf{d}$, respectively.}
    \label{dpnn_exp2}
\end{figure*}

The MNIST classification results of DPNN-S and DPNN-M under the different amounts of individual geometric errors are shown in Fig.~\ref{dpnn_exp}c and \ref{dpnn_exp}d, respectively. As the phase shift errors have a minor effect on the classification performance of DPNN, we only evaluate the phase shift error in joint systematic errors (See Fig.~\ref{dpnn_exp2}c). The MNIST classification accuracy of the baseline model for an error-free system is 96.0\% for DPNN-S and 98.6\% for DPNN-M. We implemented DAT without measuring internal states (DAT w/o IS) for DPNN-S and DAT with measuring internal states (DAT w/ IS) for DPNN-M. Here, DAT with internal states for updating SEPNs was conducted in a separable mode. For both DPNN-S and DPNN-M, the test accuracy decreases rapidly when directly deploying the \emph{in silico} trained model to the physical system, where DPNN-M has a larger decrease in classification accuracy than DPNN-S due to the accumulation of more systematic errors. The PAT method~\cite{wright:2022,spall:2022} can correct the errors to some extent but is not effective when the errors become severe and accumulate layer by layer, especially for the DPNN-M with a larger network scale. For example, one can see from Fig.~\ref{dpnn_exp}d that PAT only improves the accuracy from 25.5\% obtained by direct deployment to 66.9\% when \emph{Z-Axis} shift error is set to 1 cm and fails when \emph{XY-Plane} shift error is set to 5 degrees as it only improves the accuracy from 22.6\% to 26.3\%. By contrast, DAT outperforms PAT and dramatically eliminates the performance degradation caused by different systematic errors, making the classification accuracies comparable with and even slightly higher, e.g., for the DPNN-S model with \emph{Z-Axis} shift error, than the error-free systems. These results validate the effectiveness and robustness of DAT for training DPNN physical systems, especially demonstrating its powerful capacity to adapt to significant systematic errors from various sources in large-scale DPNN-M. Moreover, the results for FMNIST classification shown in Extended Data Fig.~2 justify the same conclusion.

We further evaluated the performance of DAT for training DPNN-S and DPNN-M under joint systematic errors, as illustrated in Fig.~\ref{dpnn_exp2}. The table in Fig.~\ref{dpnn_exp2}c lists the results of six joint systematic error configurations, where both DPNN-S and DPNN-M are assigned three configurations. We took the experimental MNIST classification accuracies of 63.9\% with the direct deployment to a physical system in~\cite{zhou:2021} as a reference to design these error configurations, with comparable or larger systematic errors as reflected in the accuracies of direct deployment in Fig.~\ref{dpnn_exp2}c. Notice that the FMNIST classification accuracy of the baseline model for an error-free system is 83.8\% for DPNN-S and 85.8\% for DPNN-M. In the joint systematic errors, DAT also achieves superior classification accuracies over PAT and restores the model performance, especially in the large-scale DPNN-M with more significant joint systematic error. PAT fails to train DPNN-M with the last joint systematic error configuration for MNIST classification, as the accuracies are even lower than the direct deployment of \emph{in silico} trained model. By contrast, DAT successfully trains the DPNN-M, which significantly outperforms PAT and improves the classification accuracy by an average of 42.1\% for MNIST and 35.5\% for FMNIST. Besides, DAT has a larger improvement over the direct deployment method with an average accuracy of 48.5\% than 32.1\% in~\cite{zhou:2021} for MNIST classification. We further showed the convergence plots during the training stage in Extended Data Fig.~3, demonstrating that DAT was more robust than PAT, especially for DPNN-M.

Fig.~\ref{dpnn_exp2}a, \ref{dpnn_exp2}b, and \ref{dpnn_exp2}d illustrate the results for DPNN-M in the task of MNIST classification with the first joint systematic configuration in Fig.~\ref{dpnn_exp2}c. Fig.~\ref{dpnn_exp2}a visualizes the network internal states of the first, third, and fifth PNN blocks and final output with the example input digit `7' from the test set. The output intensities are distributed throughout the plane for \emph{in silico} trained model and around detection regions for PAT, thus leading to an incorrect recognition category. By contrast, the intensities are concentrated in the correct detector region (the bottom left one) for DAT, and thus the example testing digit `7' can be correctly categorized. Fig.~\ref{dpnn_exp2}b illustrates the phase modulation layers $\textbf{M}_{n1}$ and $\textbf{M}_{n2}$ of the $n$-th PNN block for $n=1,4,7$. The phase modulation layers obtained by PAT and DAT are dramatically different. 
Different from PAT, which generates phase modulation layers with a relatively flat distribution of values, DAT tends to generate a drastically uneven distribution to adapt to systematic errors, according to the contrast between the yellow (near $\pi$) and blue (near $-\pi$) areas.
The confusion matrices in Fig.~\ref{dpnn_exp2}d summarize the classification results of 10,000 digits in the test set and further reveal the effectiveness of DAT as it concentrates the matched pairs of predicted labels and true labels on the main diagonal.

\subsection{Training MPNN with DAT}
\label{Experiments_MZI_onns}

\begin{figure*}[!t]
    \centering
    \includegraphics[width=1.0\textwidth]{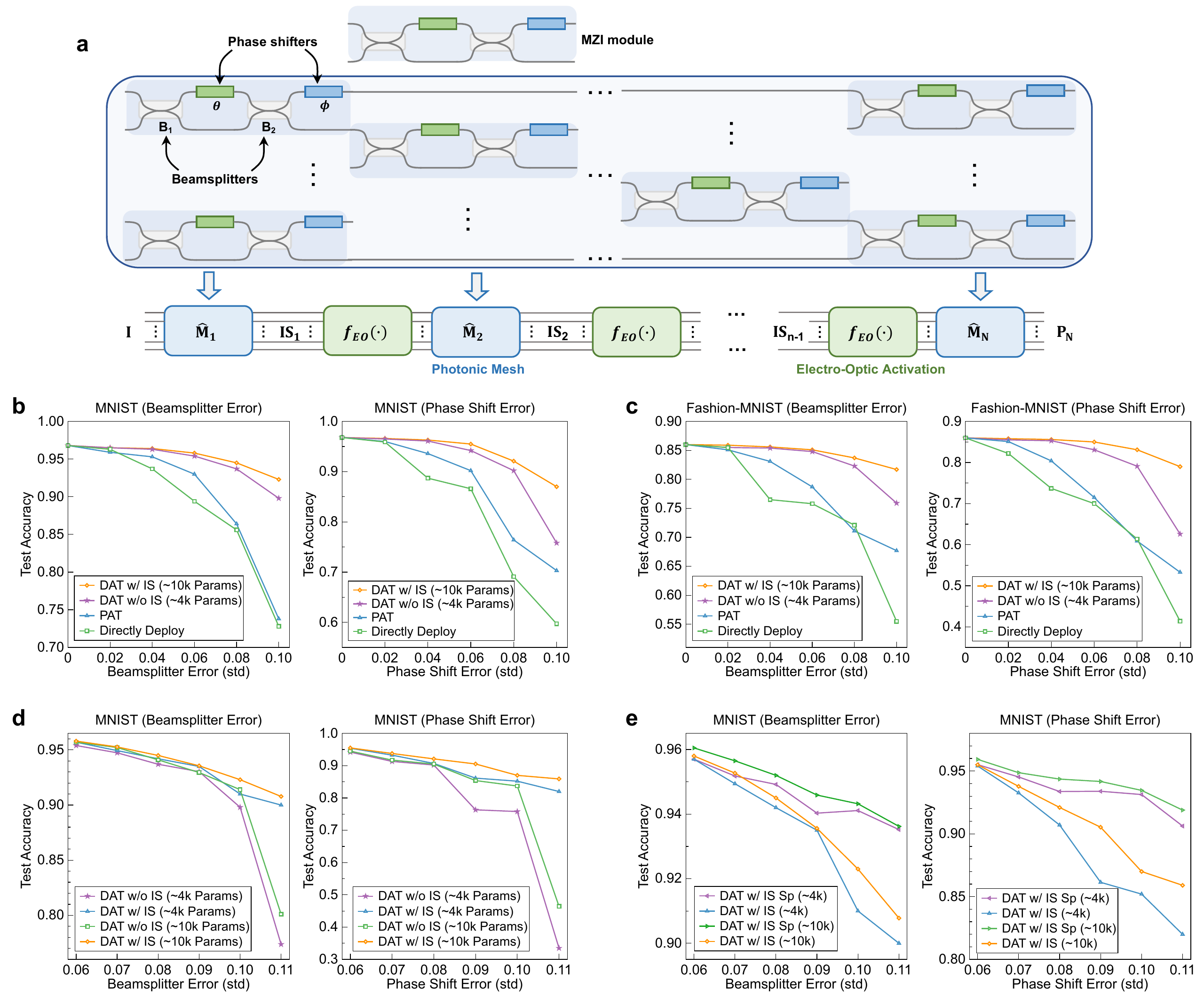}
    \caption{\textbf{Training MPNNs under two types of systematic errors for the MNIST and FMNIST classification.} \textbf{a}, Schematic illustration of the $N$-layer MPNN consisting of $N$ photonic meshes and $N-1$ optoelectronic activation units. Each photonic mesh comprises an array of MZIs with a rectangular grid (top subgraph), where each MZI consists of two beamsplitters and two phase shifters. The bottom subgraph describes the forward inference of the MPNN architecture, where $\mathrm{IS}_n$ denotes the location to obtain internal states. \textbf{b,c}, Comparing the performances of DAT with and without internal states (IS), PAT, and direct deployment of \emph{in silico} trained model on training three-layer MPNNs under beamsplitter and phase shift errors for the MNIST and FMNIST classifications. \textbf{d}, Comparisons between DAT with and without internal states under different SEPN scales. \textbf{e}, Comparisons between separable and unitary training modes for optimizing SEPNs under different SEPN scales when adopting DAT with internal states.}
    \label{mpnn_exp}
\end{figure*}

As shown in Fig.~\ref{mpnn_exp}a, the $N$-layer MPNN consists of $N$ photonic meshes and $N-1$ optoelectronic units between adjacent photonic meshes for implementing nonlinear activation functions. Each photonic mesh is constructed with the array of MZIs formed as the rectangular grid~\cite{clements:2016}. Each MZI is a two-port optical component made of two $50:50$ beamsplitters $\textbf{B}_1, \textbf{B}_2$ and two tunable single-mode phase shifters with parameters $\phi,\theta$. In the $n$-th photonic mesh, the input optical field encoded in single-mode waveguides is multiplied with a unitary matrix $\hat{\textbf{M}}_n$ realized by the $n$-th photonic mesh. The result is further processed with an optoelectronic unit with the function $f_{EO}(\cdot)$ for nonlinear processing, except for the final photonic mesh, to generate the output optical fields for the next layer. The $f_{EO}(\cdot)$ is the optoelectronic nonlinear activation function introduced in~\cite{williamson:2019} (see Methods for the formulation). The output intensity at the last photonic mesh is measured by photodetectors and used for obtaining the inference result of a task. The mathematical forward model and the training process of DAT for the MPNN are elaborated in Methods. Similar to DPNNs, all SEPN modules for MPNNs share the same complex-valued mini-UNet architecture yet are lighter than the counterparts utilized in DPNN training. Specifically, we constructed each SEPN module with two different numbers of learnable parameters, i.e., 9,648 and 3,960 parameters, to evaluate the influence of the SEPN scale on the classification performance. Compared with the standard UNet with 7,765,442 parameters, the parameter ratios of the two SEPNs are 0.124\% and 0.051\%, respectively.

\begin{figure*}[!t]
    \centering
    \includegraphics[width=1.0\textwidth]{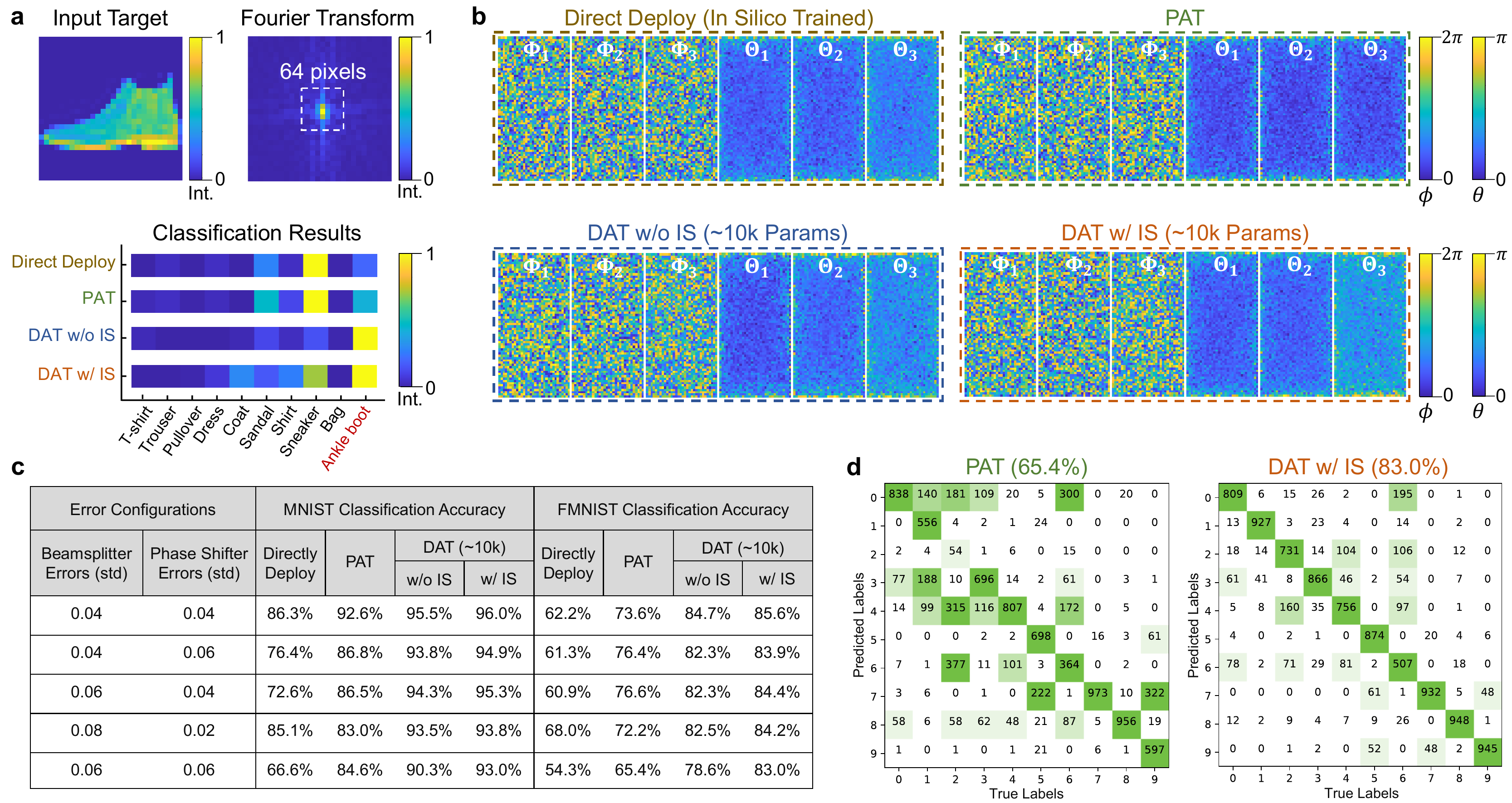}
    \caption{\textbf{Training MPNN under joint systematic errors for the MNIST and FMNIST classification.} The performances of DAT with and without using internal states are evaluated on the MPNN architectures and compared with the PAT and direct deployment of \emph{in silico} trained model under different joint systematic error configurations, as shown in Table $\textbf{c}$. The last configuration of MPNN listed in Table~$\textbf{c}$ was selected for the visualizations of an example result of fashion product `ankle boot', phase shifter values, and confusion matrices on the FMNIST classification in $\textbf{a}$, $\textbf{b}$, and $\textbf{d}$, respectively.}
    \label{mpnn_exp2}
\end{figure*}

The input data are pre-processed to facilitate the on-chip implementation of MPNNs with a limited number of input ports (see Methods and Fig.~\ref{mpnn_exp2}a). Similar to~\cite{williamson:2019}, we extracted 64 Fourier coefficients in the center region of the Fourier-space representations as the input for MNIST and FMNIST classification. To match the input dimension, each photonic mesh consists of $64\times 63/2=2016$ MZIs (see Methods and Supplementary Appendix~C) that contains 4032 beamsplitters, 2016 phase shifters with parameters $\phi$, and 2016 phase shifters with parameters $\theta$. We built the MPNN with $N=3$ for MNIST and FMINST classification, where the MPNN settings and training process are detailed in Methods. We considered two kinds of systematic errors occurring in MZIs, i.e., beamsplitter error and phase shifter error~\cite{pai:2019, pai:2019_arxiv}, caused by the imperfection of fabrications and inaccuracy of optical modulations. The beamsplitter error and phase shifter error are modeled with a normal distribution with zero mean and standard deviation of $\sigma_{bs}$ and $\sigma_{ps}$, respectively. Besides, the errors are included in all devices and share the same strengths. For example, $\sigma_{ps}=0.1$ means that the error corrupts all the 4032 phase shifters following a normal distribution with zero mean and standard deviation of 0.1.

We compare the DAT with PAT and direct deployment of \textit{in silico} trained model for the MNIST and FMNIST classification under two types of systematic errors in Fig.~\ref{mpnn_exp}b and \ref{mpnn_exp}c, respectively. The legends marked with `$\sim$10k Params' or `$\sim$4k Params' denote implementing DAT with the learnable parameters of 9,648 or 3,960 for each SEPN. All the internal states of intensities measured at the output of photonic meshes are utilized for DAT with internal states (IS), and the performance of DAT with partially measured internal states is discussed in Extended Data Fig.~4. The classification accuracy of the baseline MPNN model in an error-free system is 96.8\% for MNIST and 86.0\% for FMNIST. As shown in Fig.~\ref{mpnn_exp}b and \ref{mpnn_exp}c, DAT outperforms PAT and direct deployment even without measuring internal states and with the relatively small SEPN scale. By contrast, PAT confronts the difficulty of training, especially under large systematic errors. For example, the accuracy is 71.1\% when using PAT for FMNIST classification under $\sigma_{bs}=0.08$, while the accuracy for direct deployment is 72.1\% with the same error. At the same time, DAT with and without internal states achieve classification accuracies of 83.7\% and 82.3\%, respectively, indicating the effectiveness of SEPNs for characterizing systematic errors during the DAT training. In the individual error configurations with the largest stds listed in Fig.~\ref{mpnn_exp}b and \ref{mpnn_exp}c, DAT with internal states exceeds PAT by 16.0\% (MNIST, $\sigma_{bs}=0.1$), 8.2\% (MNIST, $\sigma_{ps}=0.1$), 9.2\% (FMNIST, $\sigma_{bs}=0.1$), 9.3\% (FMNIST, $\sigma_{ps}=0.1$), and the data for DAT with internal states increase to 18.5\%, 14.0\%, 16.7\%, and 25.7\%. The results demonstrate the superior performance and robustness of DAT for training MPNN with significant systematic errors.

We further evaluated the influence of the SEPN scale on the performance of MNIST classification under different amounts of systematic errors with the std range from 0.06 to 0.11 in Fig~\ref{mpnn_exp}d. The accuracies are close with relatively small errors (from 0.06 to 0.08), while the gap becomes evident as the std increases (from 0.09 to 0.11). With the same SEPN scale, DAT with internal states outperforms DAT without internal states, especially under larger systematic errors. Although a large SEPN scale facilitates the classification, one can find that it is not evident for beamsplitter error yet effective for phase shifter error. Compared with the SEPN with about 4k parameters, the larger one with about 10k parameters improves the test accuracy by 0.7\% and 7.17\% on average for beamsplitter and phase shifter error, respectively. Furthermore, Fig~\ref{mpnn_exp}e compares the performances between unitary and separable mode (denoted by the term `Sp' in the legends) of DAT with internal states for the MNIST classification. The separable mode surpasses the unitary mode significantly with the same SEPN scale, especially with increased std. Besides, DAT with the larger scale SEPNs improves the accuracy trivially for beamsplitter errors no matter which training mode is chosen, but it has a relatively significant improvement on performance for phase shifter errors.

The classification results of MPNN under five joint systematic error configurations with different error strengths are listed in Fig.~\ref{mpnn_exp2}c. Here, DAT was implemented with and without internal states and configured the SEPN with $\sim$10k parameters in a unitary optimization mode. DAT without internal states can adapt moderate systematic errors, e.g., it can improve the MNIST/FMNIST classification accuracy from 72.6\%/60.9\% of direct deployment to 94.3\%/82.3\% when $\sigma_{bs}=0.06$ and $\sigma_{ps}=0.04$. When corrupted by severe errors, measuring internal states can obviously improve the test accuracy, e.g., DAT with internal states exceeded DAT without internal states by 2.7\%/4.4\% for MNIST/FMNIST classification when $\sigma_{bs}=0.06$ and $\sigma_{ps}=0.06$. Meanwhile, DAT outperforms PAT by a large margin, especially under severe errors. Fig.~\ref{mpnn_exp2}a, \ref{mpnn_exp2}b, and \ref{mpnn_exp2}d visualize the results for FMNIST classification when $\sigma_{bs}=0.06$ and $\sigma_{ps}=0.06$. Fig.~\ref{dpnn_exp2}a depicts the visualization of the intensities of input and output of the example product `ankle boot'. The input is the 64-pixel values in the center region of the Fourier-space representations, and the output is the intensities on 10 photodetectors corresponding to 10 categories. The classification results demonstrate that \emph{in silico} trained model and PAT fail to classify the example to the true category (the last detector), whereas DAT suppresses the errors and successfully obtains the true classification result. Fig.~\ref{dpnn_exp2}b illustrates the phase shift values $\mathbf{\Phi}_1, \mathbf{\Phi}_2, \mathbf{\Phi}_3$ consisting of all phase shifter values of $\phi$ ranging from 0 to $2\pi$ and $\mathbf{\Theta}_1, \mathbf{\Theta}_2, \mathbf{\Theta}_3$ consisting of all phase shifter values of $\theta$ within (0, $\pi$). Fig.~\ref{mpnn_exp2}d further plots the confusion matrices representing the classification results of 10,000 products in the FMNIST test set, showing that DAT can effectively optimize the MPNN to extract the characteristics of some products that are hard to identify for PAT. For example, only 5.4\% products of `pullover' (category No.2) were correctly categorized for PAT, and the accuracies soared to 73.1\% for DAT with internal states.

\section{Discussion}\label{discussion}

In this work, we propose DAT for effectively training large-scale PNNs under significant systematic errors. The PNN numerical model, comprising the physical model and SEPNs, of the physical system, is optimized through dual backpropagation training in an end-to-end form that iteratively updates the parameters for each input training sample. Compared with the existing \textit{in situ} training methods, e.g., using extensive system measurements with layer-by-layer training process~\cite{zhou:2021} or additional hardware configurations for backward optical field propagation~\cite{hughes:2018, zhou:2020}, the DAT is a more general approach and cost-efficient for training large-scale analog PNN systems. It only requires forward inference to record output intensity with the optional internal states. For example, for the MPNN with $L$-dimensional input vector and $N$ photonic meshes in Fig.~\ref{mpnn_exp}a, the \textit{in situ} training method in \cite{hughes:2018} requires $3NL(L-1)/2$ intensity measurements of phase shifters, i.e., three times of measures for each phase shifter, to generate of backward optical fields for each training sample in addition to the output intensity. In contrast, DAT with internal states only needs to measure $N$ output intensities of photonic meshes, and DAT without internal states only requires the final network output without generating backward optical fields. Meanwhile, we have validated that DAT achieves more accurate gradient calculations than PAT for the more robust optimization of large-scale PNNs under various inherent systematic errors of varying strengths. More comparisons of the proposed DAT with the existing PNN training methods are provided in Supplementary Appendix~D.

The underlying principle of DAT for high-precision \textit{in situ} training is to transfer the additional hardware complexity to scalable algorithm complexity by introducing SEPNs during the numerical modeling. Intuitively, the parameters of SEPNs need to be proportional to the error strengths. Empirically, we found that the total parameters of the SEPN setting at the same scale with respect to the physical model can create enough fitting capacity for systematic errors. In this work, the parameter ratio between the SEPNs and physical model is approximately 1.0 for training the DPNN and MPNN with `$\sim$4k' parameters. Besides, we found that increasing the SEPN scale with the parameter ratio to approximate 2.4 for training MPNN with `$\sim$10k' parameters has considerably less influence on the performance than the training strategies, including whether measuring the internal states or not and using unitary or separable training mode. Furthermore, the learnable parameters of SEPNs with the complex-valued mini-UNet are significantly lighter than standard UNet (even 0.051\% for the ratio of parameters) is enough to dramatically eliminate the performance degradation under various errors. For the connectivity, we incorporate SEPNs into the physical model with residual connections~\cite{he:2016}, which is demonstrated to be effective in independently modeling the inherent systematic errors. Suppose the SEPNs are connected directly from the input to the output for each PNN physical layer. In that case, the SEPN needs to share the responsibility for modeling the physical computing process, resulting in the requirement of large ESPN scale and inefficiency of learning.

DAT has significant advantages in training PNNs under larger systematic errors compared with the state-of-the-art \textit{in situ} training approaches, which facilitates the training of larger network scale and mitigates the system and fabrication precision, such as the translation stages for alignments in DPNNs and the on-chip fabrications of beamsplitter and phase shifter in MPNNs. Besides, we propose the unitary and separable mode of DAT for training PNNs with or without internal states to deal with different scenarios. Generally, the separable mode is more robust with higher performance, especially for large-scale PNNs, as it refines the optimization of SEPNs without apparently increasing the computational complexity. Furthermore, although the DAT only be examined on DPNN and MPNN in this work, it's a general \textit{in situ} training paradigm that can be applied for universal PNN training or other types of AI systems with analog computing errors.

\section{Methods}

\subsection{Preprocessing of benchmarks}\label{data_preprocessing}
The two benchmark datasets for classification, i.e., MNIST and FMNIST, consist of 70,000 grayscale $28\times 28$ pixel images of 10 handwritten digits and fashion products of 10 classes, respectively. Specifically, the MNIST dataset contains digit categories from 0 to 9, while the FMNIST dataset contains product types of t-shirt, trousers, pullover, dress, coat, sandal, shirt, sneaker, bag, and ankle boots. In addition, both datasets consist of a training set of 60,000 examples and a test set of 10,000 examples.

For the DAT training of DPNNs, the images were first upscaled to the size of $100\times 100$ using bilinear interpolation to facilitate the physical fabrication and the network optimization. Then, to further satisfy the boundary condition of free-space propagation numerically implemented with the angular spectrum method~\cite{bueno:2018, lin:2018, zhou:2020}, the images were further padded with zeros to the size of $200\times 200$. Following previous works~\cite{lin:2018, zhou:2020}, we adopted coherent illumination to encode the input information. Specifically, these preprocessed images were encoded into the amplitude of the complex optical fields with zero phases under a working wavelength of 1550 nm.

For the DAT training of MPNNs, we cropped the data to reduce the input size following the steps in~\cite{williamson:2019}. Specifically, the images with an original size of $28\times 28$ were first converted into two-dimensional Fourier-space representations, then 64 pixels in the center region of the representations, i.e., $8\times 8$ Fourier coefficients closest to the center point, were extracted as input since the Fourier-space energy is mostly concentrated in the low-frequency domain located in the center region. This process is illustrated in Fig.~\ref{mpnn_exp2}a. After the date preprocessing, the 64-dimensional signals were input to the MPNNs through 64 input strip optical waveguides. The input information was compressed with the compression ratio of $64/28^2\approx 8.16\%$ in comparison to the original data. Higher compression ratios can achieve better classification performance but require more input ports and raise costs.

\subsection{DPNN settings}\label{DPNN_settings}
We built DPNN-S and DPNN-M as shown in Fig.~\ref{dpnn_exp}a and \ref{dpnn_exp}b, and evaluated the DAT training performance with two models. Following previous works~\cite{zhou:2021}, the phase modulation layers can be implemented using programmable spatial light modulators (SLMs). In our DPNNs, the pixel size of SLMs was set to 17 $\mu$m under the working wavelength of 1550 nm during the training and testing. Each phase modulation layer was well configured by packing up $200\times 200$ neurons, i.e., there are $40,000$ learnable parameters for each layer covering an area of $3.4$ mm $\times$ $3.4$ mm on the SLM. The total number of input nodes and neurons determined by the phase modulation layers is 0.12 million and 0.84 million for the DPNN-S and DPNN-M, respectively. The periphery of the phase modulation layer was zero-padded to guarantee the boundary condition of free-space diffraction during the numerical modeling with the angular spectrum method~\cite{bueno:2018, lin:2018, zhou:2020}. Besides, the distances between successive layers were fixed to 30 cm for the DPNN-S and 10 cm for the DPNN-M. Both distances for optical diffraction enable a fully connected structure according to the Huygens–Fresnel principle for calculating the maximum diffractive angle~\cite{lin:2018}. The sigmoid function was used to limit the phase modulation range of each element to $0\sim 2\pi$~\cite{lin:2018, zhou:2020}, which facilitates DPNN training and makes full use of the full-range phase modulation of SLM. All the phase modulation parameters were randomly initialized before network training.

The output plane after each PNN block utilized an optoelectronic sensor to measure the intensity of the whole optical field and also served as a nonlinear activation function between adjacent PNN blocks to provide a powerful capacity for feature extraction. In the last PNN block, the output plane contains 10 detector regions corresponding to 10 classes of digits in MNIST or products in FMNIST. Each detector region covers $22\times 22$ pixels with detector width $0.374$ mm. The classification criterion is to find the detector region with the maximal intensity by optimizing the DPNNs with a cross-entropy loss function. During the training, the intensities of the pixels in each detector region were summed up and normalized by the softmax function to generate a 10-dimensional vector. Then, the cross-entropy loss function was introduced as the task loss $L_{\mathrm{t}}$ to minimize the deviation between the generated vector and the ground truth $\textbf{T}$.

\subsection{MPNN settings}\label{MPNN_settings}
We built the MPNNs as shown in Fig.~\ref{mpnn_exp}a with $N=3$ as used in \cite{williamson:2019, pai:2019_arxiv, pai:2019}. We chose Clements scheme~\cite{clements:2016} to construct MPNNs, which was discussed detailed in \cite{pai:2019}. Each photonic mesh was connected with 64 input and output optical waveguides to match the input and output dimensions, where the 64 input optical waveguides were used to input the preprocessed data, and the other 64 output optical waveguides were used to transfer the internal states to the following optoelectronic unit except for the last photonic mesh. In the last mesh, we adopted a drop-mask to reduce the final output to 10 components to match the classification categories; thus, only 10 waveguide ports were utilized. The intensities of the 10 outputs were normalized by the softmax function and then compared with the one-hot encoding of the target vector. We utilized the cross-entropy loss function as the task loss $L_{\mathrm{t}}$ to minimize the deviation between the normalized output intensities and the correct one-hot vector and optimize the numerical model. In addition, each photonic mesh consisted of $64\times 63/2=2016$ MZIs (see Supplementary Appendix~C for more details) that contained 4032 beamsplitters, 2016 phase shifters with parameters $\phi$, and 2016 phase shifters with parameters $\theta$. At the start of the training process, each $\phi$ was initialized to a random value in $[0, 2\pi]$ following a uniform distribution, i.e., $\phi\sim U[0, 2\pi]$, and each $\theta$ was initialized following $\theta\sim U[0, \pi]$.

\subsection{Training details of DPNN}\label{training_details_DPNN}
Both the DPNN-S and DPNN-M were trained using a stochastic gradient descent algorithm, i.e., the adaptive moment estimation (Adam) optimizer~\cite{kingma:2014} with $\beta_1=0.9$ and $\beta_2=0.999$, during the \emph{in silico} training, PAT and DAT processes. With \emph{in silico} training to obtain the baseline accuracy in an error-free system, the physical model of DPNN-S was optimized for 10 epochs with a batch size of 32 and an initial learning rate $0.01$ decayed by 0.5 every epoch, while DPNN-M was trained for 50 epochs with a batch size of 128 and an initial learning rate $0.01$ decayed by 0.5 every 10 epoch due to the large scale. Besides, DPNN-S were trained for both 5 epochs with PAT and DAT, while DPNN-M were trained for 50 and 10 epochs with PAT and DAT, respectively. Except for the optimization of the physical model, there is another gradient descent step to update SEPNs for each training sample during the DAT process. The same Adam optimizer was utilized with a constant learning rate $0.001$ to minimize $L_{\mathrm{s}}$ and optimize all the SEPNs in a unitary or separable mode. All the experiments were performed on a desktop computer with Intel Xeon Gold 6226R CPU at 2.90GHz with 16 cores and an Nvidia GTX-3090Ti GPU of 24 GB graphics card memory.

\subsection{Training details of MPNN}\label{training_details_MPNN}
We adopted the same Adam optimizer as the experiments of DPNNs to train the MPNN with \emph{in silico} training, PAT, and DAT. Specifically, the physical model of the MPNN as in Fig~\ref{mpnn_exp}(a) with $N=3$ was trained for 50 epochs with a batch size of 32 and an initial learning rate $0.001$ decayed by 0.5 every 10 epochs during the \emph{in silico} training, PAT and DAT processes. The extra gradient descent step in DAT for optimizing SEPNs was implemented with an initial learning rate of $0.001$ that was decayed by 0.5 every 20 epochs. In addition, during the first 20 epochs of DAT w/o IS, SEPNs were optimized to predict the systematic errors, but the connection with the physical model was cut off during the optimization of the physical model, which meant that SEPNs did not participate in the backpropagation process and the calculation of gradients. Since the SEPNs haven't fully characterized the systematic errors in the first few epochs, it may cause unstable optimization when involved in updating the physical model. After the first 20 epochs, SEPNs were roughly trained and then reconnected with the physical model to implement a standard optimization in the last 30 epochs.

\subsection{Architecture of SEPNs}\label{Architecture_SEPN}
All SEPNs shared the same architecture as illustrated in Extended Data Fig.~1 in this work. Inspired by UNet~\cite{ronneberger:2015}, each SEPN was designed as a complex mini-UNet with hierarchically interconnected structures to extract multiscale features, while it was much simpler and lighter than UNet. To match the complex-valued computation of DPNNs and MPNNs, We adopted complex-valued weights similar to~\cite{trabelsi:2018} to construct the SEPNs. As shown in Extended Data Fig.~1, each \textbf{c}omplex-valued \textbf{conv}olution layer (CConv) is set to a size of $5\times 5$ in the DPNN and $3\times 3$ in the MPNN, followed by a \textbf{c}omplex-valued \textbf{ReLU} (CReLU) except for the last convolution layer, where the CReLU is defined as: $\mathrm{CReLU}(\textbf{x})=\mathrm{ReLU}(\mathrm{Re}\{\textbf{x}\}) + j\cdot\mathrm{ReLU}(\mathrm{Im}\{\textbf{x}\}).$ With input size of $H\times W$, successive CConvs with stride 2 (green blocks) are introduced to downscale the size to $\frac{H}{2}\times \frac{W}{2}$ and $\frac{H}{4}\times \frac{W}{4}$, while two \textbf{c}omplex-valued \textbf{t}ransposed \textbf{conv}olution layers (CTConvs) with stride 2 (yellow blocks) are utilized to upsample the size from $\frac{H}{4}\times \frac{W}{4}$ to $\frac{H}{2}\times \frac{W}{2}$, and from $\frac{H}{2}\times \frac{W}{2}$ to $H\times W$. Other CConvs plotted within blue blocks convolute input with stride 1 to maintain the scale yet may change the feature channel numbers.

The total number of learnable parameters for a SEPN can be calculated as: $k^2(4F_1+2F_1^2+4F_1F_2+2F_2^2+2F_2F_3+2F_3^2)$, where $F_1, F_2$, $F_3$ denote the numbers of feature channels, and $k$ represents the convolutional kernel size. In the experiments for DPNN-S and DPNN-M, we set $F_1=4, F_2=8, F_3=16$, and $k=5$; thus, the parameter number is 26,800. As for the MPNN, we construct two SEPNs with different scales. The lighter one was configured with $F_1=4, F_2=6, F_3=8$ and $k=3$ with a parameter number of 3,960, and $F_1, F_2, F_3, k$ for the other were set to 4, 8, 16, 3 with a parameter number of 9,648. Compared with UNet~\cite{ronneberger:2015} with a parameter number of 7,765,442, the SEPNs are lighter and can be efficiently optimized.

\subsection{Description of systematic errors}\label{description_of_systematic_errors}
We considered four types of errors that occurred in DPNN practical systems, including the phase shift error that causes biased phase modulations and the geometric errors containing \emph{Z-Axis} shift error, \emph{X-Axis} shift error, and \emph{XY-Plane} rotation error. \emph{Z-Axis} shift error denotes the propagation distance error of optical diffraction, \emph{X-Axis} shift error denotes the upper shift error of the phase modulation layers and the output plane, and \emph{XY-Plane} rotation error represents the rotation deviation of the phase modulation layers and the output plane, described using angles. In addition, we modeled the phase shift error values to be independently sampled from a normal distribution with zero mean and std $\sigma$. We include the geometric errors in a layer-by-layer manner, each with the same amount of errors. For example, setting \emph{Z-Axis} shift error to 1 cm in DPNN-S means that each of $\textbf{W}_{11}, \textbf{W}_{12}, \textbf{W}_{13}$ deviates with an extra propagation distance 1 cm. Thus, the holistic emph{Z-Axis} shift error is 3 cm. Setting \emph{XY-Plane} shift error to 1 degree in DPNN-M means that the two-phase modulation layers and the output plane rotate with 1 degree relative to the previous layer or plane for every PNN block. Thus, the holistic \emph{XY-Plane} shift error accumulates to 9 degrees.

As for the MPNN, there are mainly two types of systematic errors, i.e., beamsplitter error and phase shifter error~\cite{pai:2019, pai:2019_arxiv}. Beamsplitter error is caused by imperfect beamsplitters with split ratio errors that change the behavior of the perfect $50:50$ coupling regions, the formulation of which is described in Supplementary Eq.~C5. The phase shifter error can affect the value of $\phi$ and $\theta$, leading to uncertainties of phase modulation. Various error sources may result in the slight performance deviance of phase shifters in the photonic meshes, such as thermal crosstalk or environmental drift~\cite{pai:2019}. Following the same assumptions made for DPNN, the beamsplitter and phase shifter error values were independently sampled from a normal distribution with zero mean and std $\sigma_{bs}$/$\sigma_{ps}$, and the errors are included in all devices and share the same strength. For example, setting $\sigma_{ps}=0.1$ means that all the 4,032 phase shifters are corrupted by such a phase shifter error for the MPNN with $N=3$.

\subsection{Separable mode for training SEPNs}\label{separable_mode}
For a $N$-layer PNN with input $\mathbf{I}$, we can obtain the observations $\{\mathbf{P}_n\}_{n=1}^N$ from the physical system by measuring internal states and final output, with which the counterparts $\{\bar{\mathbf{S}}_n\}_{n=1}^N$ from the numerical model can be extracted. Here, $\{\bar{\mathbf{S}}_n\}_{n=1}^N$ are not equal to the $\{\mathbf{S}_n\}_{n=1}^N$ obtained by unitary inference of the numerical model with the same initial input $\textbf{I}$. The $\{\mathbf{S}_n\}_{n=1}^N$ obtained by unitary inference are not appropriate to be the extracted internal states and output to approximate the targets $\{\mathbf{P}_n\}_{n=1}^N$ in the separable training mode, as the inputs of the $n$-th group for obtaining $\mathbf{S}_n$ and $\mathbf{P}_n$ are mismatched. To address this issue, we propose to extract $\{\bar{\mathbf{S}}_n\}_{n=1}^N$ with a separable inference of the numerical model. Specifically, for the $n$-th group, the input is replaced by the corresponding practical intensity, then $\bar{\mathbf{S}}_n$ is generated by the inference of the group with the replaced input. This process is repeated $N$ times to produce $\{\bar{\mathbf{S}}_n\}_{n=1}^N$. The extraction of $\{\bar{\mathbf{S}}_n\}_{n=1}^N$ for the DPNN-M is illustrated in Supplementary Appendix E and Supplementary Fig.~S1. Nevertheless, $\{\mathbf{S}_n\}_{n=1}^N$ is still indispensable for the optimization of the physical model, although it is not used for the separable training of SEPNs. In addition, although the process is based on the assumption that all the internal states are measured, it can be easily extended to the scenario with partial intensity measurements.

For optimizing the SEPNs in a separable mode, the PNN numerical model can be separated into $N$ groups, where the $n$-th group corresponds to the paired data $(\mathbf{P}_n, \bar{\mathbf{S}}_n)$ and governs SEPNs within the group. For example, the DPNN-M can be divided into seven groups where the $n$-th group governs three SEPNs within the $n$-th PNN block. We denote the parameters of SEPNs in the $n$-th group by $\Lambda_n$ and the similarity loss function for the $n$-th group by $L_{\mathrm{s,n}}$, we have: $L_{\mathrm{s,n}}(\textbf{P}_n, \vert \bar{\textbf{S}}_n \vert ^2)=l_{\mathrm{mse}}(\textbf{P}_n, \vert \bar{\textbf{S}}_n \vert ^2)=\|\textbf{P}_n -  \vert \bar{\textbf{S}}_n\vert ^2 \|_2^2$. For each training sample, the parameters of the PNN physical model are fixed, and the gradients of $L_{\mathrm{s,n}}$ with respect to $\Lambda_n$ are calculated during the backpropagation to iteratively and separably optimize the SEPNs within the $n$-th group for $n\in[1, N]$. The pseudo-code of training DPNN in the separable mode is provided in Supplementary Algorithm~S2.

\subsection{DAT with internal states for DPNN}\label{methods_dpnn}
We elaborate on DAT with internal states for training DPNN by cascading multiple blocks in Fig. 2a, termed DPNN-C, to show the principle, which could be easily extended to DPNN-S and DPNN-M. The standard backpropagation process for DPNN-C is established in Supplementary Appendix~B. Extended Data Fig.~5 depicts the procedure of DAT with internal states for training the $n$-th PNN block, where $\mathrm{SEPN}_{nk}$ for $k=1,2,3$ represents the SEPN attached to corresponding diffractive propagation layers, $\textbf{W}'_{ni}$ and $\textbf{W}_{ni}$ for $i=1,2,3$ denote the ideal and practical diffractive weight matrices, $\textbf{S}_{n}$ and $\textbf{P}_{n}$ denote the simulated and practical output intensity of the $n$-the block, $\textbf{M}'_{n1}, \textbf{M}'_{n2}$ and $\textbf{M}_{n1}, \textbf{M}_{n2}$ represent the ideal and practical phase modulation matrices, respectively. Specifically, $\mathbf{M}'_{nk}=\mathrm{diag}(e^{2\pi j \mathbf{\Phi}_{nk}})$ for $k=1, 2$ denotes the diagonalization of the vectorized phase modulation layer with coefficient $\mathbf{\Phi}_{nk}$, and $\mathbf{M}_{nk}=\mathrm{diag}(e^{2\pi j (\mathbf{\Phi}_{nk}+\epsilon_{nk})})$, where $j$ denotes the imaginary unit and $\epsilon_{nk}$ denotes the phase shift error. Mathematically, the forward propagation of the $n$-th block for DPNN-C physical system with $N$ blocks can be formulated as follows based on the Rayleigh-Sommerfeld diffraction principle,
\begin{equation}\label{physical_prop_DPNN}
\begin{aligned}
\textbf{U}_n &= \textbf{W}_{n3}\textbf{M}_{n2}\textbf{W}_{n2}\textbf{M}_{n1}\textbf{W}_{n1}\textbf{P}_{n-1}, \\
\textbf{P}_n& =\vert\textbf{U}_n\vert^2,
\end{aligned}
\end{equation}
and the corresponding forward numerical model with SEPNs can be formulated as:
\begin{equation}\label{numerical_prop_DPNN}
\begin{aligned}
\textbf{U}'_{n1} &= \textbf{M}'_{n1}\left[\mathcal{N}_{n1}(\textbf{W}'_{n1}\textbf{O}_{n-1}) + \textbf{W}'_{n1}\textbf{O}_{n-1}\right], \\
\textbf{U}'_{n2} &= \textbf{M}'_{n2}\left[\mathcal{N}_{n2}(\textbf{W}'_{n2}\textbf{U}'_{n1}) + \textbf{W}'_{n2}\textbf{U}'_{n1}\right], \\
\textbf{S}_{n} &= \mathcal{N}_{n3}(\textbf{W}'_{n3}\textbf{U}'_{n2}) + \textbf{W}'_{n3}\textbf{U}'_{n3}, \\
\textbf{O}_n& =\vert\mathbf{S}_n\vert^2,
\end{aligned}
\end{equation}
where $\mathbf{U}_{n}, \mathbf{U}'_{n1}, \mathbf{U}'_{n2}$ represent the vectorized complex optical fields; $\mathbf{O}_n$ denotes the intensity of $\textbf{S}_{n}$; $\mathcal{N}_{n1}, \mathcal{N}_{n2}, \mathcal{N}_{n3}$ denote the functions expressed by the SEPNs; $\mathbf{P}_0=\mathbf{O}_0=\mathbf{I}$ denote the initial input. The formulation is based on the residual connections~\cite{he:2016} for incorporating SEPNs into the physical model.

Four steps of DAT are repeated over all training samples to minimize the loss functions until convergence for obtaining the numerical model and physical parameters, i.e., the phase modulation matrices $\textbf{M}_{ni}$ for $i=1,2,3$, $1\leq n \leq N$. We elaborate on the steps with one training sample as follows:

First, the optically encoded $I$ is input to the physical system to perform the forward inference. In this pass, we obtain the internal states $\textbf{P}_{n}$ for $n\in[1, N-1]$ and final output intensity $\textbf{P}_{N}$. The blue dotted arrows in Extended Data Fig.~5 describe this step for the $n$-the PNN block, corresponding to Eq.~\eqref{physical_prop_DPNN}.

Second, the same training sample $I$ is digitally encoded and input to the numerical model to extract the internal states and final observation $\textbf{S}_{n}$ for $n\in[1, N]$. The yellow dotted arrows in Extended Data Fig.~5 describe this step for the $n$-the PNN block, corresponding to Eq.~\eqref{numerical_prop_DPNN}.

Third, we optimize the SEPNs' parameters $\Lambda$ by minimizing the similarity loss function in the unitary mode as Eq.~\eqref{mse_loss} or separable mode described in Methods~\ref{separable_mode} and Supplementary Appendix~E. By the way, the similarity loss function simplifies to $L_{\mathrm{s}}(\textbf{P}, \vert \textbf{S} \vert ^2) = \|\textbf{P}_N -  \vert \textbf{S}_N\vert ^2 \|_2^2$ for DAT if without internal states. The gradients of the similarity loss function with respect to $\Lambda$ are calculated via the backpropagation to optimize the SEPNs, while the parameters of the physical model are fixed. The red dotted arrows in Extended Data Fig.~5 describe this step for the $n$-the PNN block.

Fourth, we implement state fusion to obtain new internal states and the final observation for generating the new gradients of the task loss with respect to the phase modulation matrices. Specifically, for any $n\in[1,N]$, $\textbf{P}_{n}$ can contribute to the replacement of $\vert\textbf{S}_{n}\vert^2$ with $\textbf{P}_{n}$, and the fusion of $\textbf{S}_{n}$ and $\textbf{P}_{n}$ using the fusion function $F_n(\textbf{P}_n, \textbf{S}_n) = \sqrt{\textbf{P}_n} \exp(j\mathbf{\Phi}_{\textbf{S}_n})\}_{n=1}^N$. The new internal states and observation are utilized to calculate the gradients to optimize the physical parameters of the PNN numerical model by minimizing the task loss $L_\mathrm{t}$ as Eq.~\eqref{task_loss}. Therefore, the gradients of $L_\mathrm{t}$ with respect to $\mathbf{\Phi}_{nk}$ for $k=1,2$ are derived as:
\begin{equation}
\frac{\partial L_{\mathrm{t}}}{\partial \mathbf{\Phi}_{n1}}=2\mathrm{Re}\{\frac{\partial L_{\mathrm{t}}}{\partial \mathbf{S}_n}\frac{\partial \mathbf{S}_n}{\partial \mathbf{M}'_{n1}}\frac{\partial \mathbf{M}'_{n1}}{\partial \mathbf{\Phi}_{n1}}\}, ~ \frac{\partial L_{\mathrm{t}}}{\partial \mathbf{\Phi}_{n2}}=2\mathrm{Re}\{\frac{\partial L_{\mathrm{t}}}{\partial \mathbf{S}_n}\frac{\partial \mathbf{S}_n}{\partial \mathbf{M}'_{n2}}\frac{\partial \mathbf{M}'_{n2}}{\partial \mathbf{\Phi}_{n2}}\},
\end{equation}
where 
\begin{equation}\label{gradient_DPNN}
\frac{\partial L_{\mathrm{t}}}{\partial \mathbf{S}_n} = 
\begin{cases}
\frac{\partial L_{\mathrm{t}}}{\partial \mathbf{O}_N} \odot (\mathbf{S}_N)^*,& n=N,\\
2\mathrm{Re}\{\frac{\partial L_{\mathrm{t}}}{\partial \mathbf{S}_{n+1}}\frac{\partial \mathbf{S}_{n+1}}{\partial \mathbf{O}_{n}}\}\odot(\mathbf{S}_n)^*, & 1\leq n \leq N-1;
\end{cases}
\end{equation}
$\odot$ is element-wise multiplication; $*$ is conjugation of the optical field; $\mathrm{Re}\{\cdot\}$ means reserving the real part of the optical field; the form of $\partial L_{\mathrm{t}}/\partial \mathbf{O}_N$ is related to the task loss function.
We omit the detailed formulations of $\frac{\partial \mathbf{S}_n}{\partial \mathbf{\Phi}_{nk}}$ and $\frac{\partial \mathbf{S}_{n+1}}{\partial \mathbf{O}_{n}}$ as the forms are complicated when adopting SEPNs. Nevertheless, they can be easily deduced with reference to the standard propagation of DPNN-C in Supplementary Appendix~B. Then, one gradient descent step is performed to optimize the phase modulation coefficients using the calculated gradients while the parameters of SEPNs are fixed. The green dotted arrows in Extended Data Fig.~5 describe this step for the $n$-the PNN block.

\subsection{DAT without internal states for MPNN}\label{methods_mpnn}
We elaborate on DAT without internal states for the MPNN illustrated in Fig.~\ref{mpnn_exp}a. The standard backpropagation was established in Supplementary Appendix~C. Extended Data Fig.~6 illustrates the procedure of DAT without internal states for training the MPNN, where $\textbf{I}\in\mathbb{C}^{L}$ denotes the input complex optical filed, $\hat{\textbf {M}}'_n$ and $\hat{\textbf {M}}_n$ represent the ideal and practical transformation matrix of the $n$-th photonic mesh that consisting of $L(L-1)/2$ embedded MZIs, and $\textbf{Z}'_n$ and $\textbf{Z}_n$ for $1\leq n \leq N$ represent the simulated and practical output of the $n$-th photonic mesh, respectively. Mathematically, the forward propagation of the MPNN physical system can be described as:
\begin{equation}
\begin{aligned}
\textbf{Z}_1&=\hat{\textbf {M}}_{1}\textbf{I},\\
\textbf{Z}_n &= \hat{\textbf {M}}_n f_{EO}(\textbf{Z}_{n-1}),~2\leq n\leq N,
\end{aligned}
\end{equation}
where $f_{EO}(\cdot)$ is the optoelectronic nonlinear activation function introduced in~\cite{williamson:2019} that can be formulated as $f_{EO}(\mathbf{Z})=j\sqrt{1-\alpha}\exp[ -j(\beta\vert\mathbf{Z}\vert^2+\gamma)]\cos(\beta\vert\mathbf{Z}\vert^2+\gamma)\mathbf{Z}$, and $\alpha, \beta, \gamma$ are constants related to configurations of the optoelectronic unit. The corresponding forward numerical model of the MPNN can be formulated as:
\begin{equation}
\begin{aligned}
\textbf{Z}'_1&=\mathcal{N}_{1}(\hat{\textbf {M}}'_{1}\textbf{I})+\hat{\textbf {M}}'_{1}\textbf{I}, \\
\textbf{Z}'_n &= \mathcal{N}_{n}(\hat{\textbf {M}}'_n f_{EO}(\textbf{Z}'_{n-1}))+\hat{\textbf {M}}'_n f_{EO}(\textbf{Z}'_{n-1}),~2\leq n\leq N, 
\end{aligned}
\end{equation}
where $\mathcal{N}_{n}$ denotes the function expressed by SEPN$_n$ incorporated into the physical model with residual connections~\cite{he:2016}.

Four steps for DAT without internal states are repeated over all training samples to minimize the loss functions and optimize the physical model, i.e., the phase coefficients $\mathbf{\Theta}_n$ and $\mathbf{\Phi}_n$ of the $n$-th photonic mesh for $1\leq n\leq N$, where each $\mathbf{\Theta}_n$ or $\mathbf{\Phi}_n$ contain $L(L-1)/2$ coefficients. We detail the steps for one training sample as follows:

First, $\textbf{I}$ is optically encoded and inputted to the physical system through waveguide ports to implement a forward inference, with which we measure the final output intensity $\textbf{P}_N=\vert\textbf{Z}_N\vert^2$. The blue dotted arrows in Extended Data Fig.~6 describe this step.

Second, $\textbf{I}$ is digitally encoded and input to the numerical model for inference. To be consistent with $\textbf{P}_N$, we only extract the complex optical field $\textbf{S}_N=\textbf{Z}'_N$. The yellow dotted arrows in Extended Data Fig.~6 describe this step.

Third, we simultaneously optimize all the SEPNs by minimizing the similarity loss function $L_{\mathrm{s}}(\textbf{P}, \vert\textbf{S}\vert^2) = \|\textbf{P}_N-\vert \textbf{S}_N\vert^2\|_2^2$. The gradients of $L_{\mathrm{s}}$ with respect to the parameters of SEPNs are calculated during the backpropagation to optimize the SEPNs in the unitary mode for one step, while the parameters of the physical model are fixed.
The red dotted arrows in Extended Data Fig.~6 describe this step.

Fourth, we fuse $\textbf{P}_{N}$ and $\textbf{S}_{N}$ to obtain $\sqrt{\textbf{P}_{N}} \exp(j\mathbf{\Phi}_{\textbf{S}_{N}})$ to replace $\textbf{S}_{N}$, and directly replace the simulated intensity $\vert\textbf{S}_{N}\vert^2$ by $\textbf{P}_{N}$.
The new states and the replaced output are utilized to calculate the gradients of the task loss $L_{\mathrm{t}}$ as Eq.~\eqref{task_loss} with respect to the phase coefficients $\mathbf{\Theta}, \mathbf{\Phi}$ of all MZIs via backpropagation through the numerical model. Specifically, we have for $n\in[1, N]$ that
\begin{equation}
\frac{\partial L_{\mathrm{t}}}{\partial \mathbf{\Theta}_{n}} = 
2\mathrm{Re}\left\{\left(\frac{\partial L_{\mathrm{t}}}{\partial \vert\mathbf{Z}'_N\vert^2} \odot (\mathbf{Z}'_N)^*\right)^T \frac{\partial \mathbf{Z}'_N}{\partial \hat{\textbf {M}}'_{n}}\frac{\partial \hat{\textbf {M}}'_{n}}{\partial \mathbf{\Theta}_{n}} \right\}, 
\end{equation}
where $\odot$, $*$ and $\mathrm{Re}\{\cdot\}$ are defined in Eq~\eqref{gradient_DPNN}; the forms of $\partial L_{\mathrm{t}}/\partial \vert\mathbf{Z}'_N\vert^2$ and $\partial \hat{\textbf {M}}'_{n}/\partial \mathbf{\Theta}_n$ are related to the task loss function and the scheme to construct MPNN, respectively. Similarly, the gradient of $L_{\mathrm{t}}$ with respect to $\mathbf{\Phi}_n$ can be easily deduced. The detail form of $\frac{\partial \mathbf{Z}'_N}{\partial \hat{\textbf {M}}'_{n}}$ is omitted due to the complexity of its formulation with SEPNs but can be deduced with reference to the standard propagation of the MPNN in Supplementary Appendix~C. Then, one gradient descent step is implemented to optimize the phase coefficients, while the parameters of SEPNs are fixed. The green dotted arrows in Extended Data Fig.~6 describe this step.

\bmhead*{Data availability}
All data needed to evaluate the conclusions in the paper are present in the paper and/or the Supplementary Materials.

\bmhead*{Code availability}
The code for the construction of MPNN is available at \url{https://github.com/solgaardlab/neurophox} and the code for training MPNN with DAT will be released soon.

\bibliography{main}

\clearpage 
\bmhead*{Acknowledgements}
This work is supported by the National Key Research and Development Program of China (No. 2021ZD0109902) and the National Natural Science Foundation of China (No. 62275139 and No. 61932022).

\bmhead*{Author contributions}
X.L. and H.X. initiated and supervised the project. X. L., Z.Z., and Z.D. conceived the research. X. L. designed the methods. Z.Z. and Z.D. implemented the algorithm and conducted experiments. Z.Z., Z.D., H.C., R.Y., S.G., and H.Z. processed the data. X.L., Z.Z., Z.D., H.C., and R.Y. analyzed and interpreted the results. All authors prepared the manuscript and discussed the research.

\bmhead*{Competing interests}
The authors declare no competing interests.

\clearpage
\begin{appendices}
\renewcommand{\figurename}{\footnotesize{\bf Extended Data Fig.}}

\begin{figure*}[!ht]
    \centering
    \includegraphics[width=1.0\textwidth]{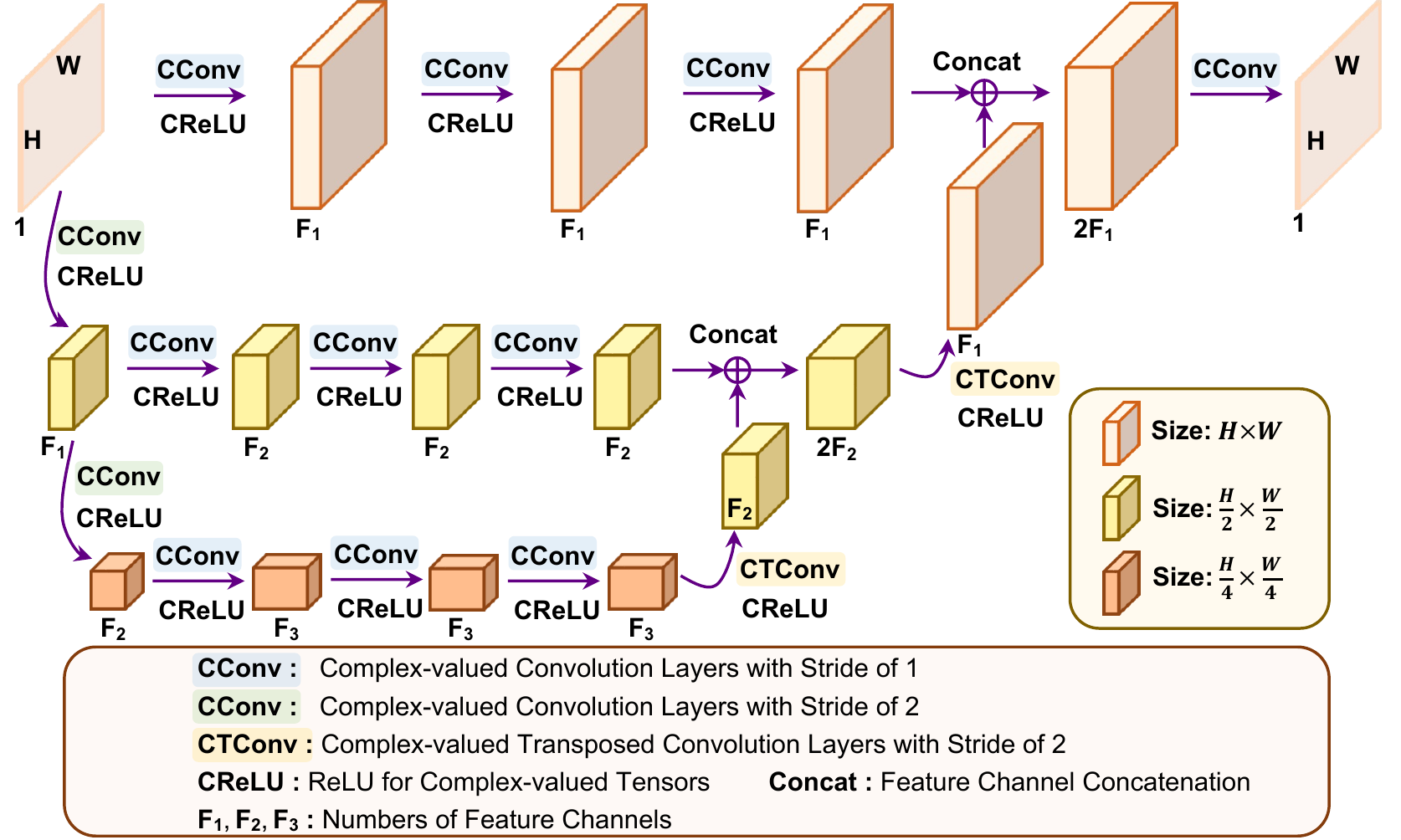}
    \caption{\textbf{Architecture of the SEPN constructed with a complex-valued mini-UNet.} See Methods for the detailed description.}
    \label{SEPN_arch}
\end{figure*}

\clearpage
\begin{figure*}[!ht]
\centering
\includegraphics[width=1.0\textwidth]{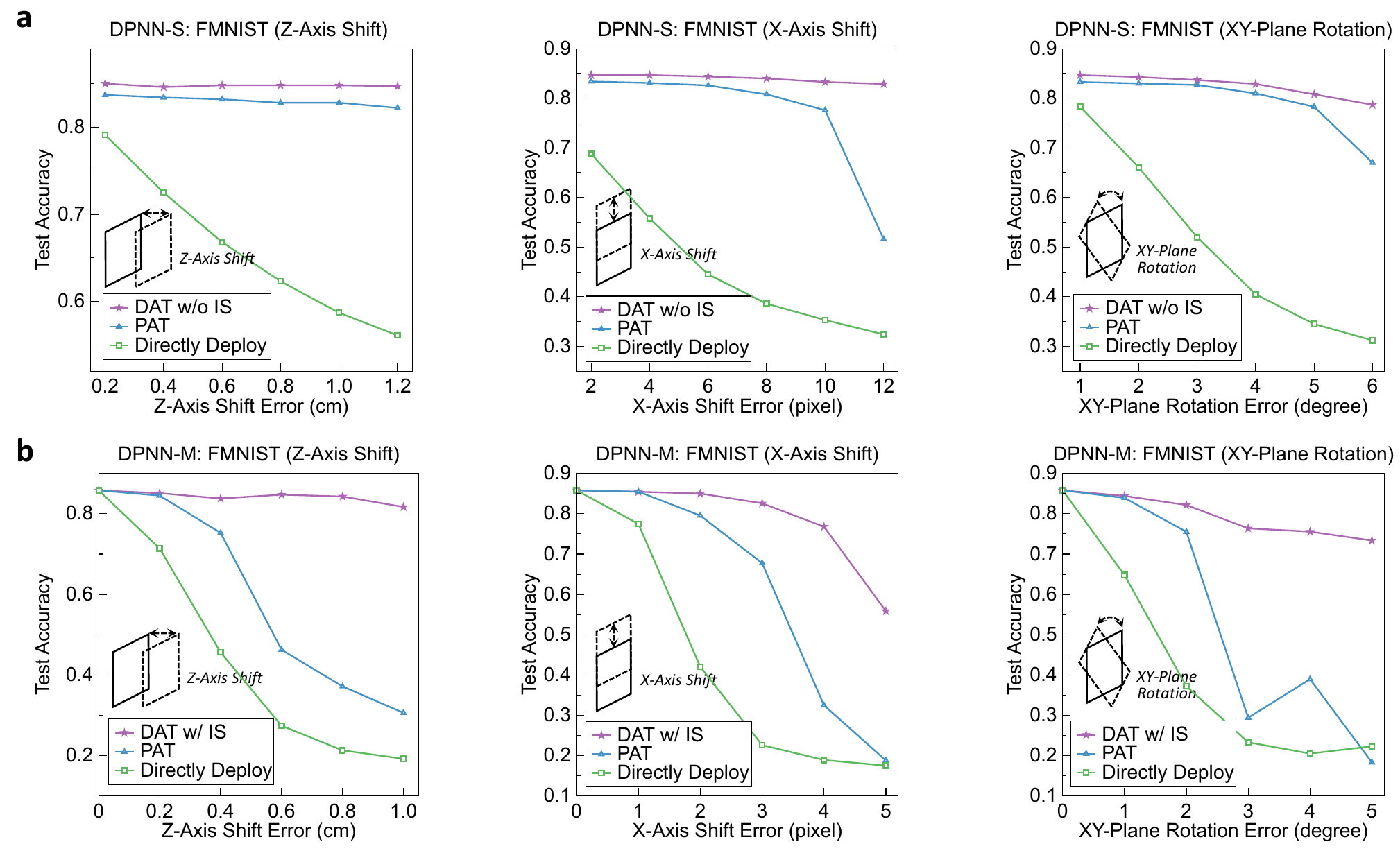}
\caption{\textbf{Training DPNNs under three types of systematic errors for the FMNIST classification.} The performances of DAT for DPNN-S ($\textbf{a}$) and DPNN-M ($\textbf{b}$) are compared with the PAT and direct deployment of \textit{in silico} trained models under different amounts of systematic errors. The DPNN-S and DPNN-M are trained without and with the internal states (IS), respectively.}
\label{DPNN_FMNIST}
\end{figure*}

\clearpage
\begin{figure*}[!ht]
\centering
\includegraphics[width=1.0\textwidth]{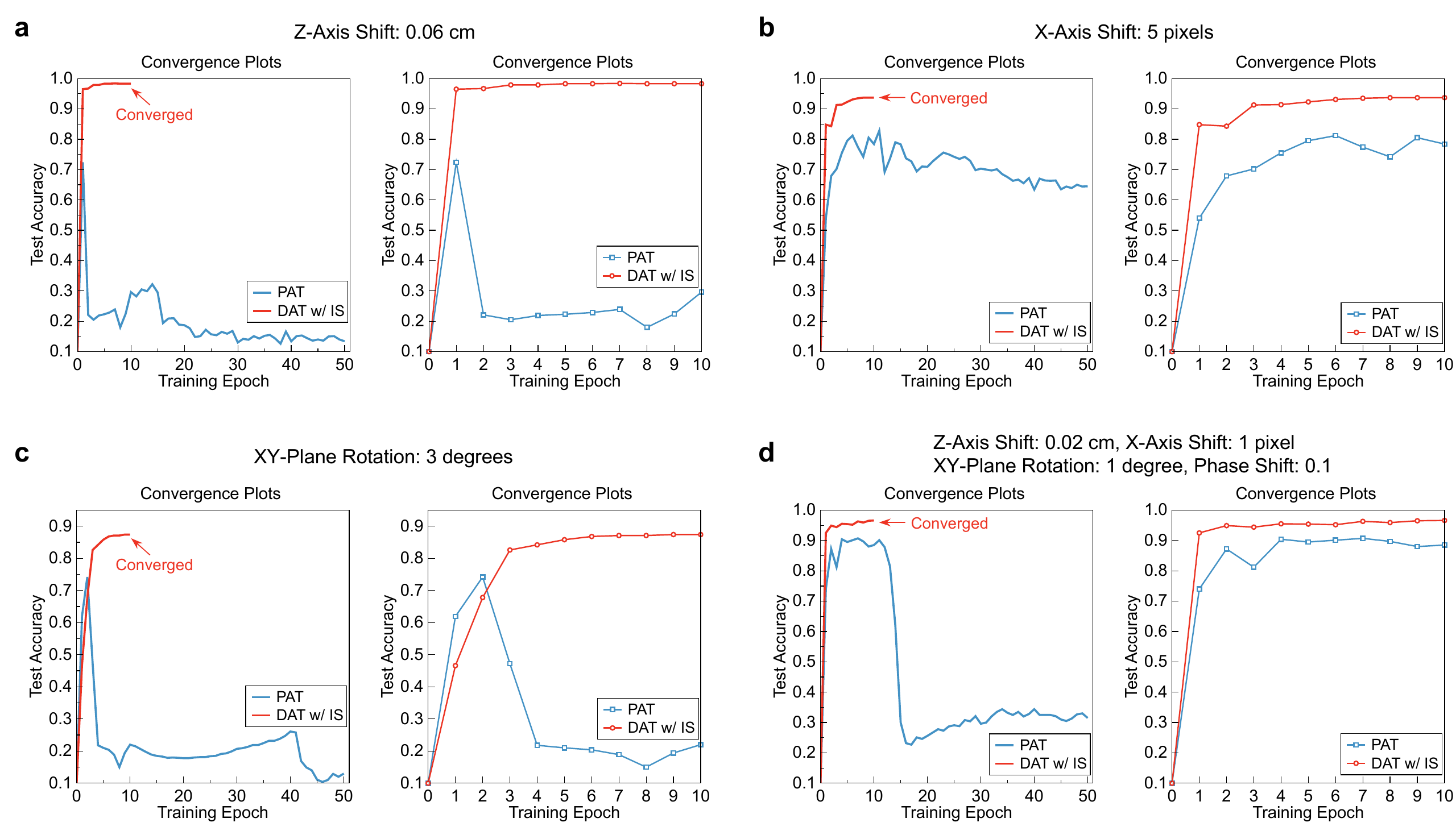}
\caption{\textbf{Convergence plots of the DPNN-M evaluated on the MNIST blind-test dataset during the training process.} Each subfigure consists of convergence plots with 50 (total training epochs for PAT, left) and 10 (total training epochs for DAT, right) epochs, where \textbf{a,b,c} represent the training process under the individual errors and \textbf{d} under the joint errors, with the error configurations shown above the subfigures. DAT outperforms PAT with a more robust training process for the optimization of the DPNN-M.}
\label{Convergence_plots}
\end{figure*}

\clearpage
\begin{figure*}[!ht]
\centering
\includegraphics[width=1.0\textwidth]{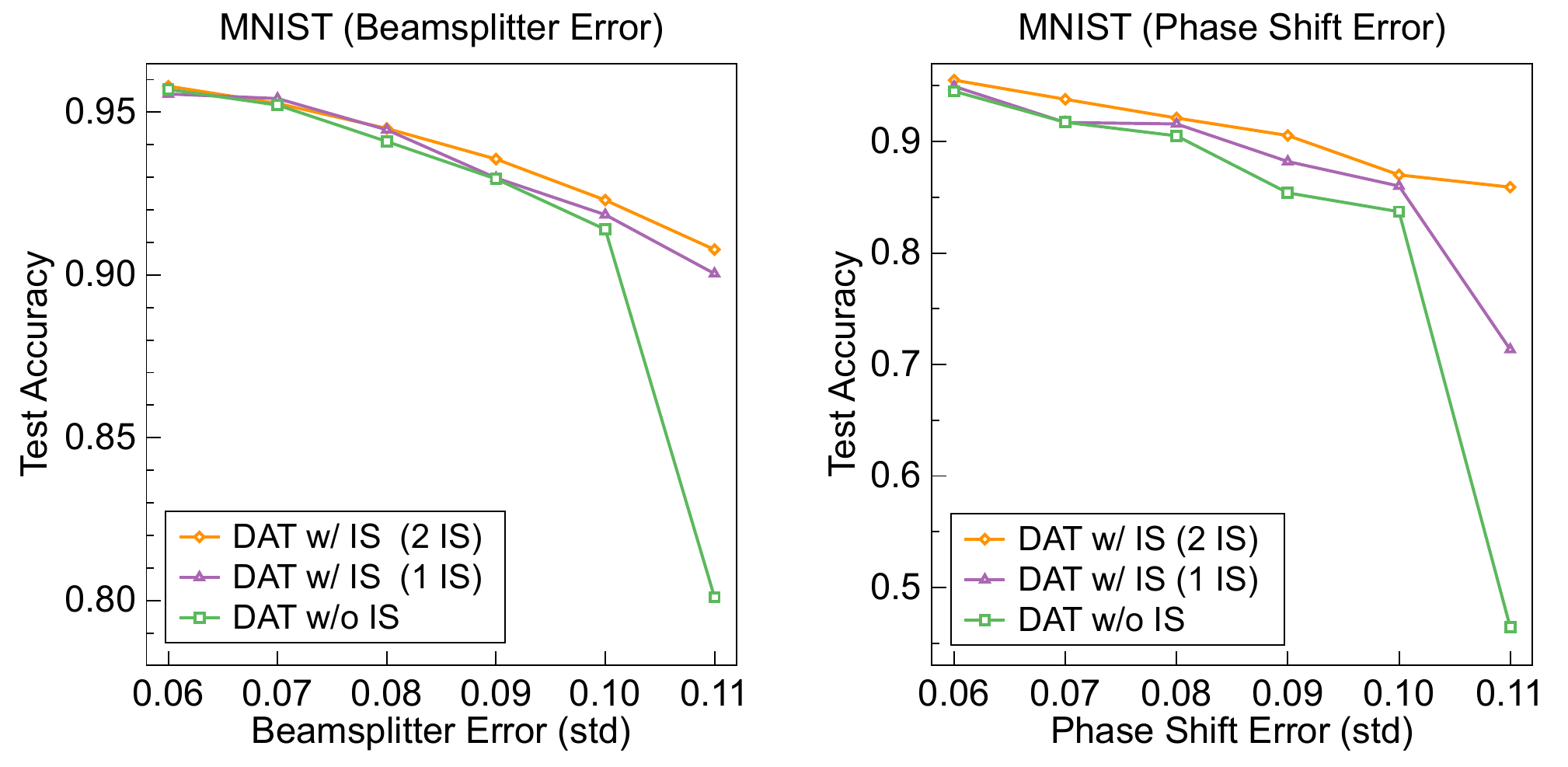}
\caption{\textbf{Comparisons of DAT performances with all, partial, and without internal states for the 3-layer MPNN in the task of MNIST classification.} The DAT methods are implemented with each SEPN parameter of 9,648 in the unitary mode. The performance of DAT with all internal states $\mathbf{P}_1, \mathbf{P}_2$ (2 IS), one internal state $\mathbf{P}_2$ (1 IS), and without internal states are evaluated. The classification accuracy improves with more measurements of internal states, especially under severe systematic errors.}
\label{partial_IS}
\end{figure*}

\clearpage
\begin{figure*}[!ht]
    \centering
    \includegraphics[width=1.0\textwidth]{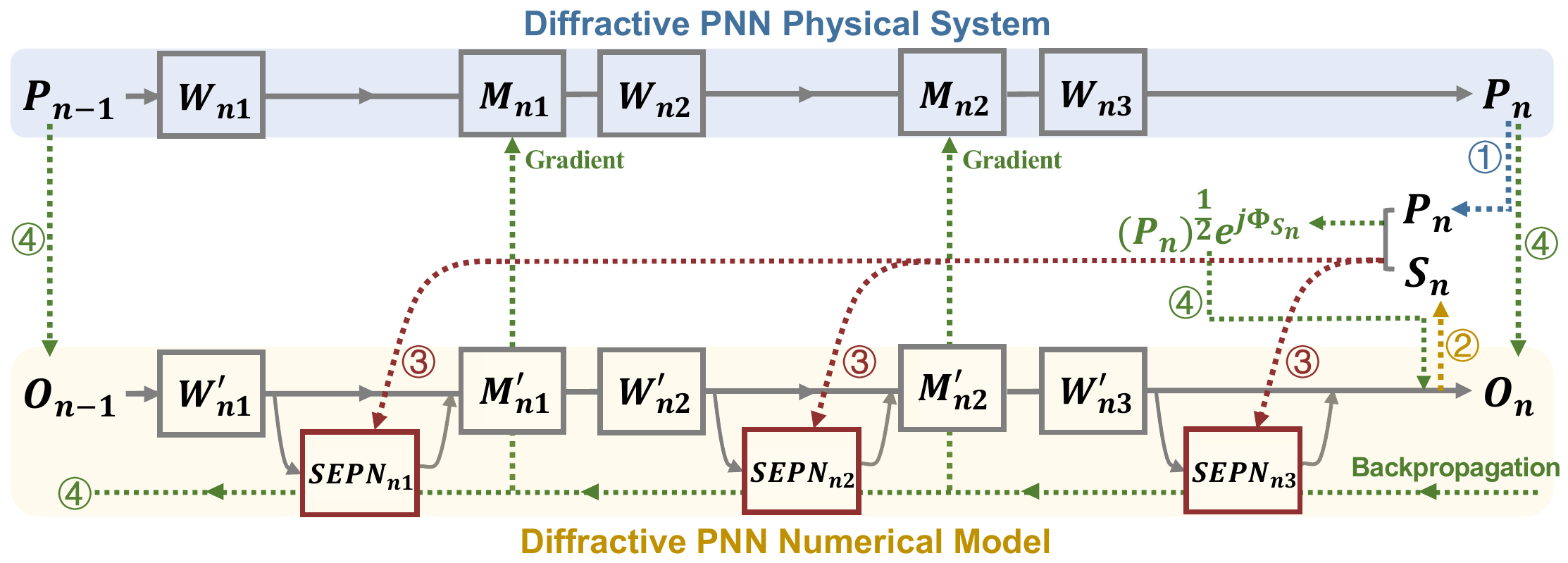}
    \caption{\textbf{Procedure of DAT with internal states for DPNN in the $n$-th PNN block.} The flow charts with blue and yellow backgrounds denote the forward inferences in the physical system and the numerical model, respectively. Four steps of DAT with internal states, labeled using dotted arrows with four different colors, are repeated over all training samples to minimize the loss functions until convergence for obtaining the numerical model and physical parameters for the system, i.e., the phase modulation matrices $\textbf{M}_{ni}$ for $i=1,2,3$, $1\leq n \leq N$. See Methods for the detailed description.}
    \label{DAT_for_DPNN}
\end{figure*}

\clearpage
\begin{figure*}[!ht]
    \centering
    \includegraphics[width=1.0\textwidth]{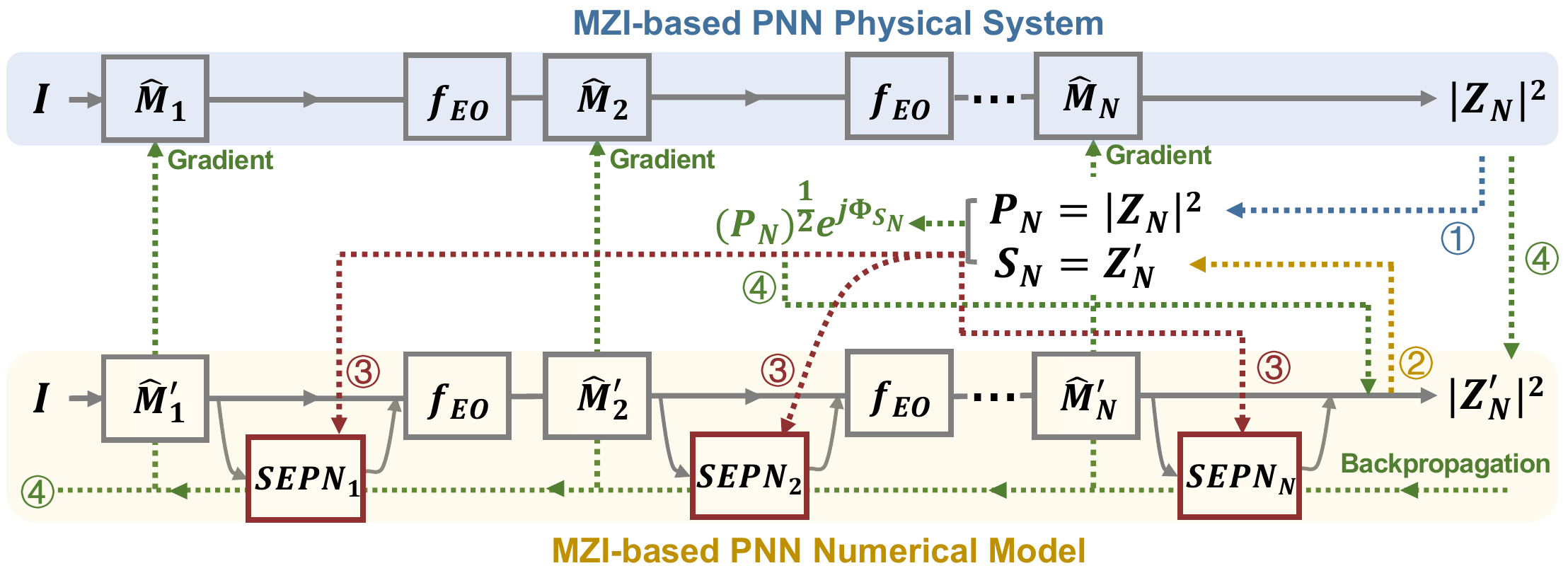}
    \caption{\textbf{Procedure of DAT without internal states for the MPNN.} The flow charts with blue and yellow backgrounds denote the forward inferences in the physical system and the numerical model, respectively. Four steps of DAT without internal states, labeled using dotted arrows with four different colors, are repeated over all training samples to minimize the loss functions and optimize the physical model, i.e., the phase coefficients $\mathbf{\Theta}_n$ and $\mathbf{\Phi}_n$ of the $n$-th photonic mesh for $1\leq n\leq N$. 
    See Methods for the detailed description.}
    \label{DAT_for_MPNN}
\end{figure*}

\end{appendices}
\end{document}